\documentclass[final]{cvpr}

\usepackage{bm}
\usepackage{textcomp} 

\usepackage{multirow}
\usepackage{subcaption}
\usepackage{amssymb} 
\usepackage{pifont} 
\usepackage{makecell} 

\usepackage{xcolor}
\definecolor{azure(colorwheel)}{rgb}{0.0, 0.5, 1.0}

\newcommand{\D}{\mathsf{D}}
\newcommand{\G}{\mathsf{G}}
\newcommand{\F}{\mathsf{F}_{\bm\theta}}

\newcommand{\ours}{\textcolor{azure(colorwheel)}{(ours)}}

\newcommand{\R}{\mathbb{R}}

\newcommand{\apref}[1]{\ref{#1}}

\usepackage{times}
\usepackage{epsfig}
\usepackage{graphicx}
\usepackage{amsmath}
\usepackage{amssymb}
\usepackage[ruled,vlined]{algorithm2e}

\usepackage[pagebackref=true,breaklinks=true,colorlinks,bookmarks=false]{hyperref}

\def\cvprPaperID{8135} 
\def\confYear{CVPR 2021}

\begin{document}

\title{Adversarial Generation of Continuous Images}


\author{Ivan Skorokhodov$^1$, Savva Ignatyev$^2$, Mohamed Elhoseiny$^1$\\
$^1$King Abdullah University of Science and Technology (KAUST)\\
$^2$Skolkovo Institute of Science and Technology\\
{\tt\small iskorokhodov@gmail.com ~~ savvaignatiev@gmail.com ~~ mohamed.elhoseiny@kaust.edu.sa} \\
}



\maketitle

\pagestyle{empty} 

\begin{abstract}
In most existing learning systems, images are typically viewed as 2D pixel arrays. However, in another paradigm gaining popularity, a 2D image is represented as an implicit neural representation (INR) --- an MLP that predicts an RGB pixel value given its $(x,y)$ coordinate.
In this paper, we propose two novel architectural techniques for building INR-based image decoders: factorized multiplicative modulation and multi-scale INRs, and use them to build a state-of-the-art continuous image GAN.
Previous attempts to adapt INRs for image generation were limited to MNIST-like datasets and do not scale to complex real-world data.
Our proposed INR-GAN architecture improves the performance of continuous image generators by several times, greatly reducing the gap between continuous image GANs and pixel-based ones.
Apart from that, we explore several exciting properties of the INR-based decoders, like out-of-the-box superresolution, meaningful image-space interpolation, accelerated inference of low-resolution images, an ability to extrapolate outside of image boundaries, and strong geometric prior.
The project page is located at \href{https://universome.github.io/inr-gan}{https://universome.github.io/inr-gan}.
\end{abstract}

\section{Introduction}\label{sec:intro}

In deep learning, images are typically represented as 2D arrays of pixels.
However, there is another paradigm which views an image as a continuous function $F( \bm p) = \bm{v}$ that maps a pixel's 2D coordinate $\bm p = (x,y) \in \R^2$ to its RGB value $\bm{v} = (r, g, b) \in \R^3$.
The advantage of such a representation is that it gives a true continuous version of the underlying 2D signal instead of its cropped quantized counterpart like pixel-based representations do.
In practice, we almost never know the underlying function $F(\bm p)$ and thus have to work with its approximations.
The most popular and expressive way to approximate $F(\bm p)$ is through a neural network $\F$ ~\cite{FourierINR, SIREN}.
It is called an \textit{implicit neural representation (INR)} and is especially popular in 3D deep learning where working with voxels directly (i.e. pixels defined on a 3D grid) is too expensive \cite{DeepMeta, IM_NET, DeepSDF, Occupancy_networks}.

\begin{figure}
\centering
\includegraphics[width=\linewidth]{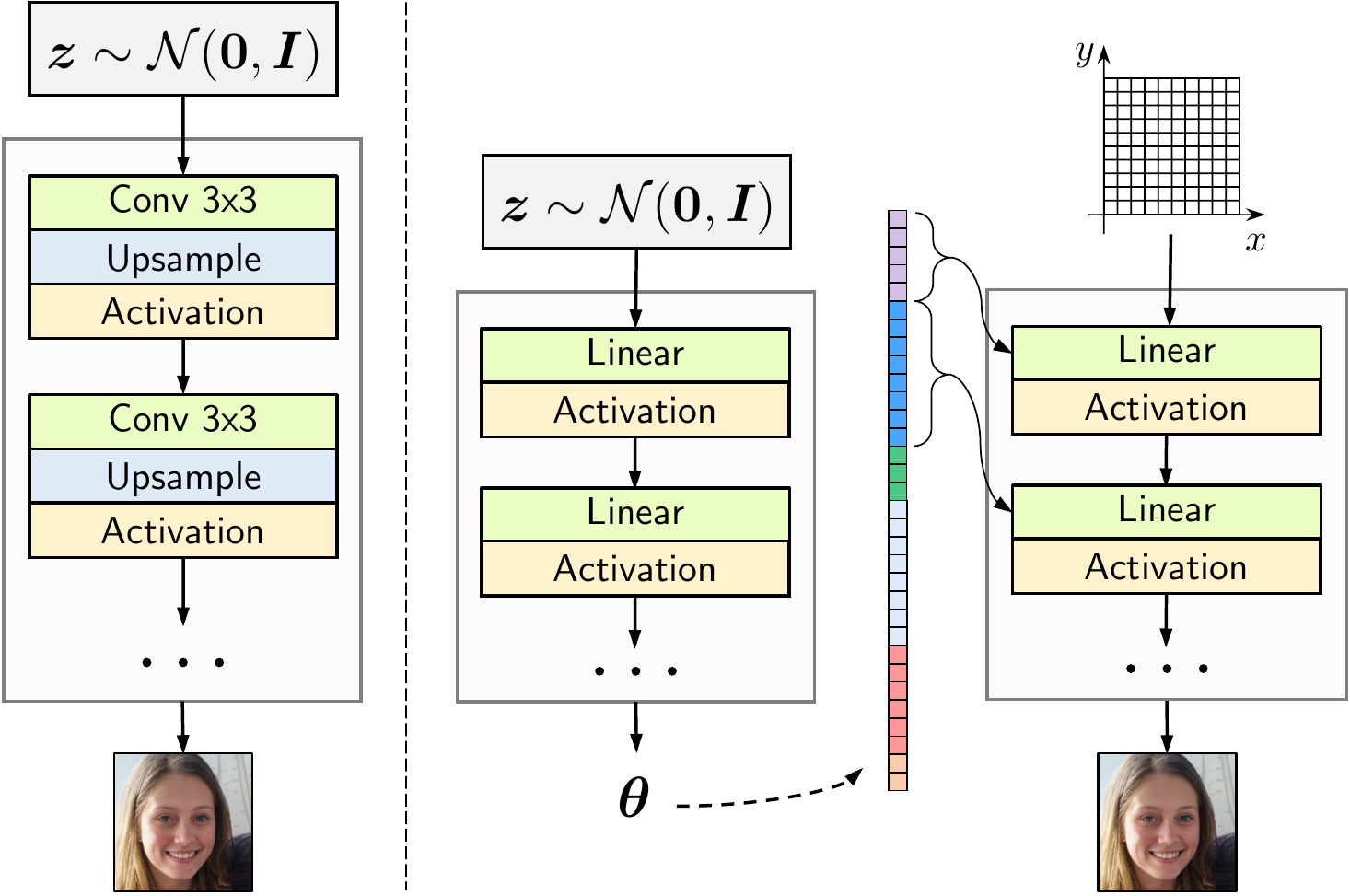}
\caption{Comparison between a traditional convolutional generator (left) and an INR-based one (right). A traditional generator directly generates a pixel-based image representation given its latent code $\bm z$. The INR-based one produces parameters of an MLP. The corresponding pixel-based representation is obtained by evaluating the INR at each coordinate location $(x,y)$ of a specified grid.}
\label{fig:inr-gan-architecture}
\end{figure}


Building such a decoder has two severe difficulties: 1) since it is a hypernetwork, i.e. a network that produces parameters for another network \cite{Hypernetworks}, it is unstable to train and requires too many parameters in general \cite{Hypernetworks_init}; and 2) it is too costly to evaluate INRs for a dense high-resolution coordinates grid limiting their application to low-resolution images only.
To alleviate these issues, we design two principled architectural techniques: \textit{factorized multiplicative modulation (FMM)} layer for hypernetworks and \textit{multi-scale} INRs.
We use these techniques to build \textit{INR-GAN}: a state-of-the-art continuous image generator that generates pictures in their INR representations.
Previous attempts to build such a model were only conducted on small MNIST-like datasets \cite{IM_NET, SpatialVAE, DeepMeta} and do not scale to complex real-world data.
In our case, we managed to achieve FID \cite{TTUR_FID} scores of 5.09, 4.96 and 16.32 on LSUN Churches $256^2$, LSUN Bedrooms $256^2$ and FFHQ $1024^2$, respectively, greatly reducing the gap between continuous image GANs and their pixel-based analogs.
In their contemporary work, \cite{CIPS} achieved even better results by employing a large-scale INR-based decoder with learnable coordinate embeddings.

\begin{figure}
\centering
\includegraphics[width=\linewidth]{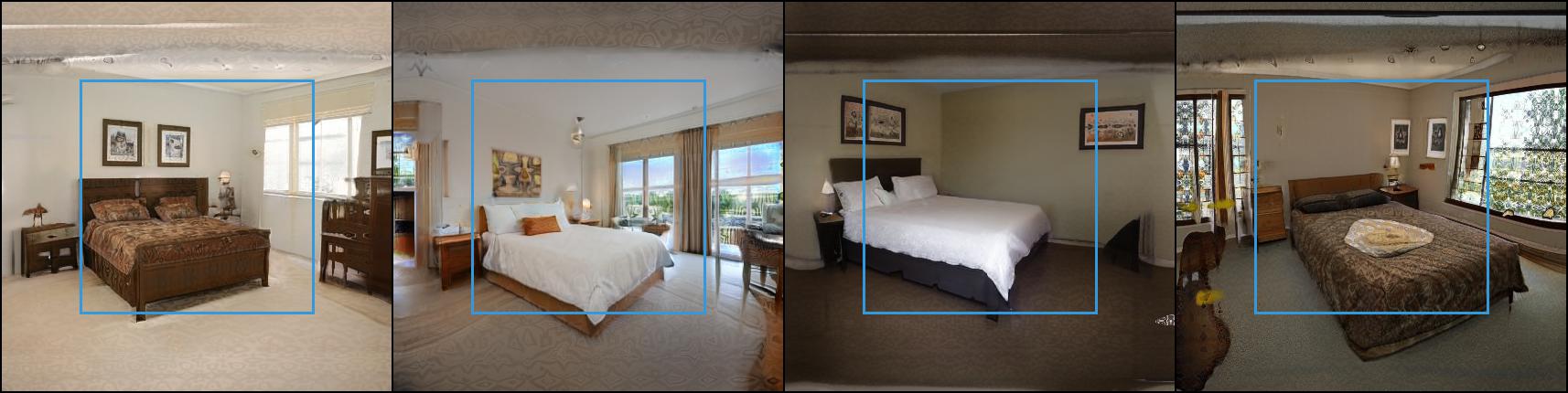}
\caption{\textbf{Extrapolating outside of image boundaries}. After training our INR-based GAN on LSUN $256^2$, we tried to evaluate it on a wider grid. During training, we used coordinates from $[0, 1]^2$ square, and here we evaluate it on coordinates from $[-0.3, 1.3]^2$ square. To our surprise, the model can produce meaningful extrapolation and generate the picture beyond the coordinates area it was trained on. Blue bounding box denotes $[0, 1]^2$ coordinates area --- an image area the model was trained on. Very similar results were previously shown by \cite{COCO_GAN}.}
\label{fig:oob-generalization}
\end{figure}

\begin{figure}
    \centering
    \begin{subfigure}[b]{0.09\linewidth}
        \centering
        \frame{\includegraphics[width=\textwidth]{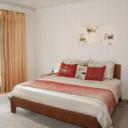}}
    \end{subfigure}
    \hfill
    \begin{subfigure}[b]{0.9\linewidth}
        \centering
        \includegraphics[width=\textwidth]{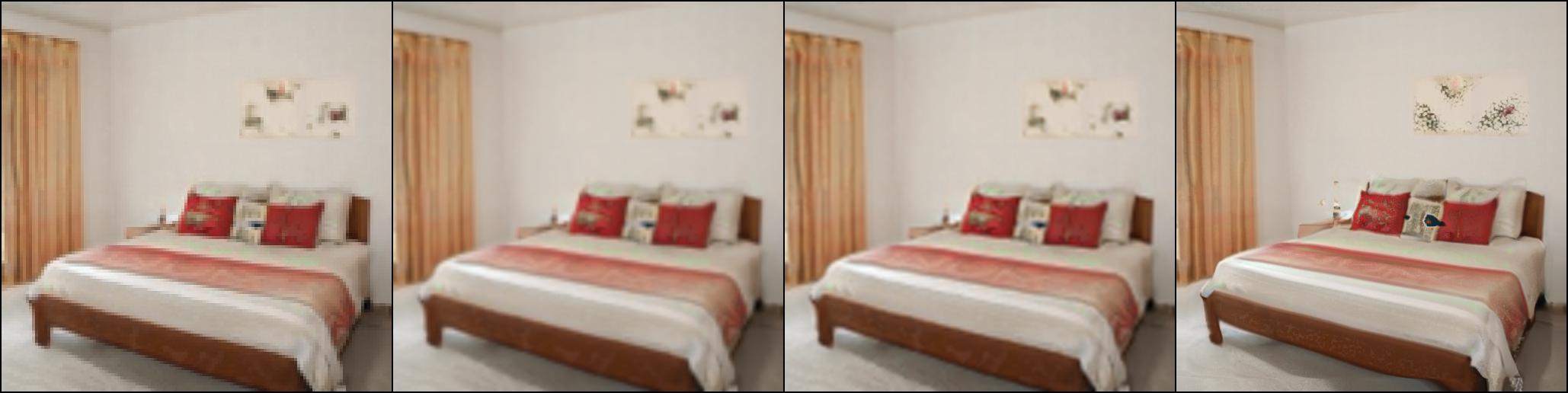}
    \end{subfigure}
    \begin{subfigure}[b]{0.09\linewidth}
        \centering
        \frame{\includegraphics[width=\textwidth]{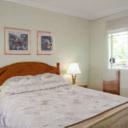}}
    \end{subfigure}
    \hfill
    \begin{subfigure}[b]{0.9\linewidth}
        \centering
        \includegraphics[width=\textwidth]{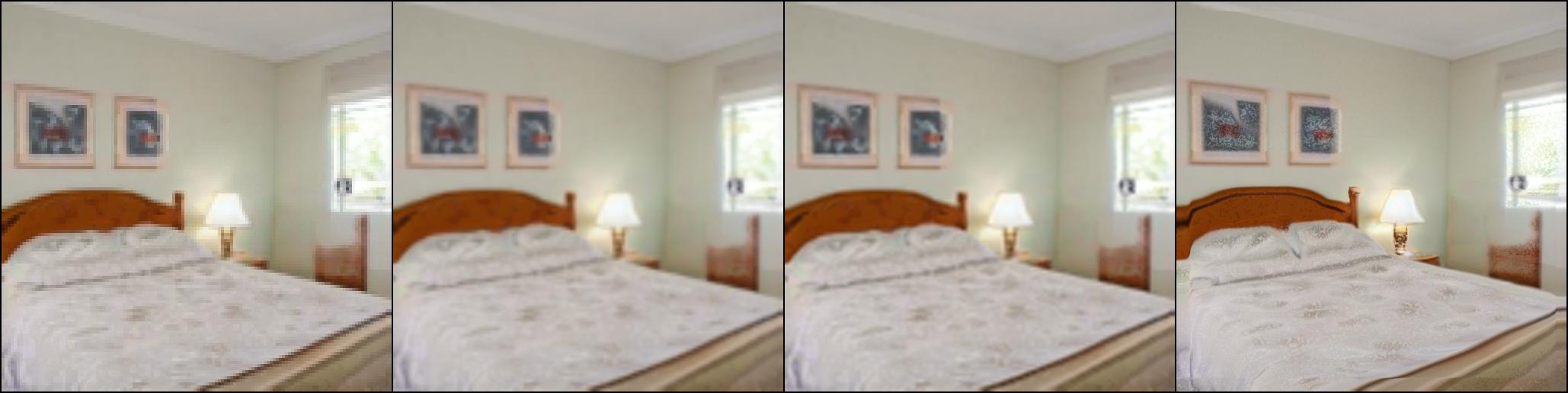}
    \end{subfigure}
    \begin{subfigure}[b]{0.09\linewidth}
        \centering
        \frame{\includegraphics[width=\textwidth]{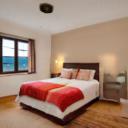}}
    \end{subfigure}
    \hfill
    \begin{subfigure}[b]{0.9\linewidth}
        \centering
        \includegraphics[width=\textwidth]{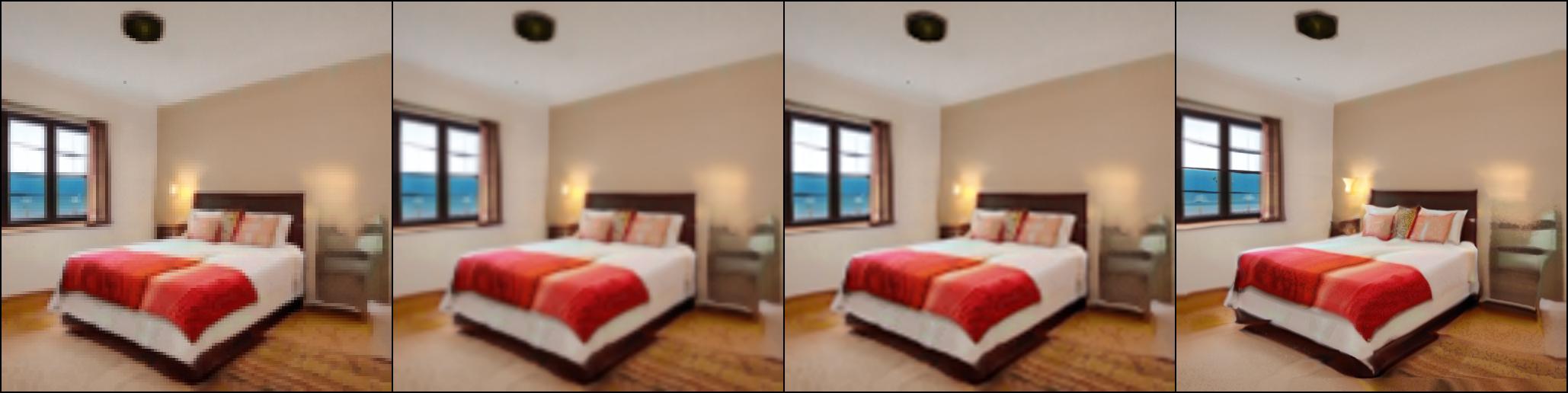}
    \end{subfigure}
    \begin{subfigure}[b]{0.09\linewidth}
        \hfill
    \end{subfigure}
    \begin{subfigure}[b]{0.9\linewidth}
        \centering
        \captionsetup[subfigure]{labelformat=empty}
        \begin{subfigure}[b]{0.22\linewidth}
            \caption{\centering Nearest (19.24)}
        \end{subfigure}
        \hfill
        \begin{subfigure}[b]{0.22\linewidth}
            \caption{\centering Bilinear (29.51)}
        \end{subfigure}
        \hfill
        \begin{subfigure}[b]{0.22\linewidth}
            \caption{\centering Bicubic (30.3)}
        \end{subfigure}
        \hfill
        \begin{subfigure}[b]{0.24\linewidth}
            \caption{\centering INR-based (11.81)}
        \end{subfigure}
    \end{subfigure}
    \caption{\textbf{Out-of-the-box superresolution properties of the INR-based decoder}. We train our INR-GAN on LSUN $128 \times 128$ and perform upsampling at test time to $256 \times 256$ resolution with either classical interpolation techniques (first 3 columns) or with ``natural'' INR-based upsampling by evaluating $\F$ on a denser coordinates grid. INR-based superresolution is much crisper and obtained without any additional supervision. Numbers in parentheses denote the corresponding UpsampledFID scores.}
    \label{fig:superresolution}
\end{figure}

\begin{figure}
    \centering
    \begin{subfigure}[b]{0.45\linewidth}
        \centering
        \frame{\includegraphics[width=0.9\linewidth]{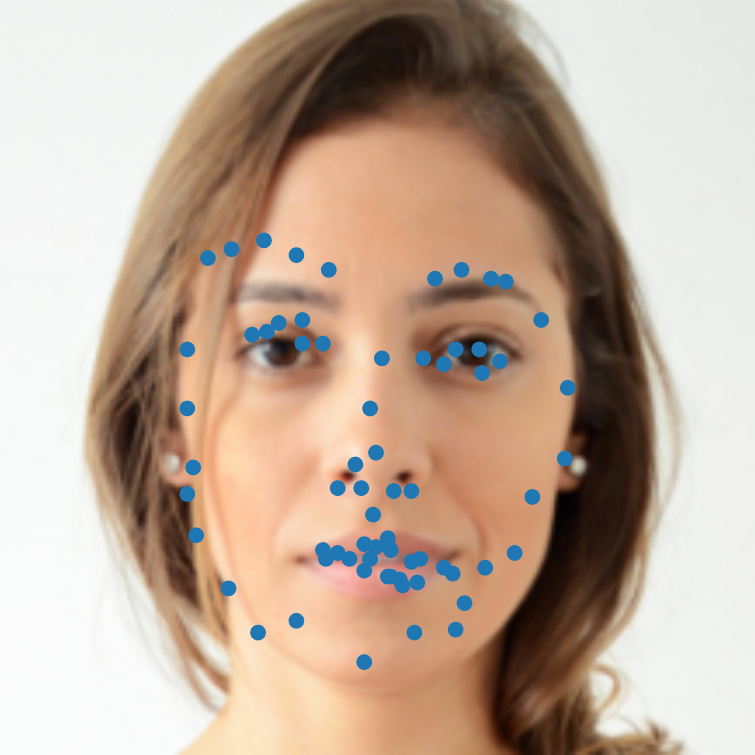}}
        \caption{StyleGAN2}
    \end{subfigure}
    \hfill
    \begin{subfigure}[b]{0.45\linewidth}
        \centering
        \frame{\includegraphics[width=0.9\linewidth]{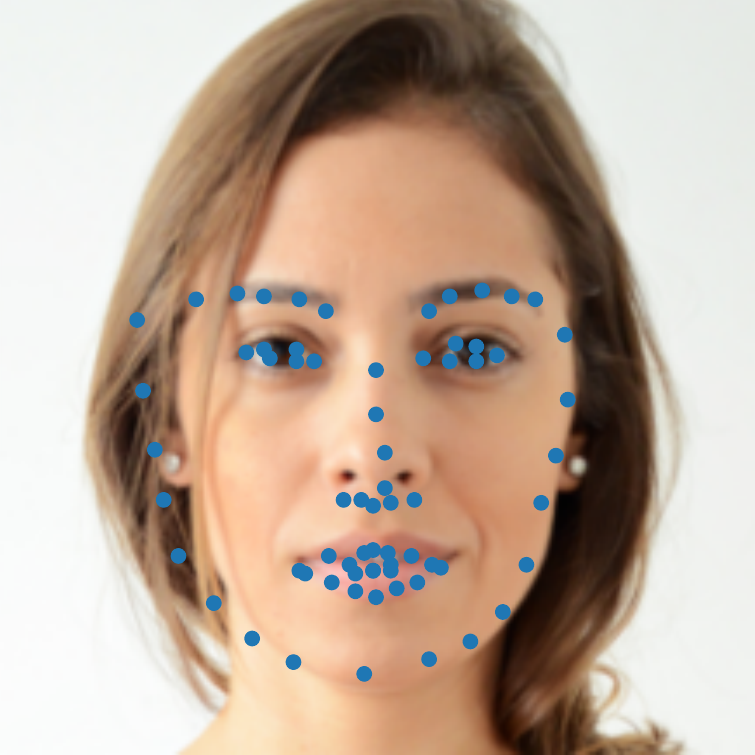}}
        \caption{INR-GAN}
    \end{subfigure}
    \caption{\textbf{Predicting keypoints from latent codes.} We train a linear model to predict face keypoints directly from the corresponding latent codes. Its performance shows how much geometric information is contained in a latent code and how accessible it is. For this benchmark, our model easily outperforms StyleGAN2 despite it being trained with the additional PPL loss \cite{StyleGAN2} that forces better latent codes conditioning in the decoder. That demonstrates better \textit{geometric prior} of our model. The corresponding scores are presented in Table~\ref{table:keypoints-prediction}.}
    \label{fig:keypoints-prediction}
\end{figure}

In our paper, we also shed light on many interesting properties of the INR-based decoders:
\begin{itemize}
    \item \textit{Extrapolating outside of image boundaries} (see ~\figref{fig:oob-generalization}): an ability to generate a ``zoomed-out'' version of an image without being trained explicitly to do this.
    \item \textit{Geometric prior} (see ~\figref{fig:keypoints-prediction}): better encoding of geometric properties of a dataset in the latent space.
    \item \textit{Accelerated low-resolution inference} (see ~\figref{fig:fast-lowres-generation}): an ability to generate a low-resolution version of an image in shorter time than an image of the corresponding training-time resolution.
    \item Meaningful image space interpolation (see ~\figref{fig:inr-space-interpolation}).
    \item \textit{Out-of-the-box superresolution} (see ~\figref{fig:superresolution}): an ability to produce a higher-resolution version of an image without being trained for this task at all.
\end{itemize}
We emphasize that these features come naturally to INR-based decoders and do not require any additional training.

To summarize, our contributions are the following:
\begin{enumerate}
    \item We propose a novel factorized multiplicative modulation (FMM) layer for hypernetworks. It makes it possible to generate INRs with a large number of parameters and stabilizes hypernetwork training. 
    \item We propose a novel \textit{multi-scale} INR architecture. It makes it possible to represent high-resolution images in the INR-based form in a very efficient way.
    \item Using the above two techniques we build INR-GAN: an INR-based GAN model that outperforms existing continuous image generators by several times on large real-world datasets.
    \item We explore several exciting properties of INR-based decoders, like out-of-the-box superresolution, meaningful image-space interpolation, accelerated inference of low-resolution images, an ability to extrapolate outside of image boundaries, and geometric prior.
\end{enumerate}

\section{Related work}

\textbf{Implicit Neural Representations}.
The original idea of augmenting neural networks with coordinates information was proposed in CPPN \cite{CPPN} which is a neuroevolution-based model that is trained to represent a 2D image.
After that, there were several works that use coordinates as an additional source of information to neural networks \cite{CoordConv, Location_augmentation_for_CNNs, SOLOv2, Pix2Pose, HybridPose}, but the largest popularity of implicit neural representations (INRs) is observed in 3D deep learning, where it provides a cheap and continuous way to represent a 3D shape compared to mesh/voxel/pointcloud-based ones \cite{Occupancy_networks, DeepSDF, DeepMeta, Structured_implicit_functions, Deep_Structured_implicit_functions, Deepvoxels}.
They have also been used for other tasks, like representing textures \cite{Texture_fields}, 3D shapes flow \cite{Occupancy_flow}, scenes \cite{NeRF, SRN}, audios and differential equations \cite{SIREN}, human grasps \cite{Grasping_Field} and other information \cite{Implicit_representation_layers, 3D_Unet}.
Occupancy Networks \cite{Occupancy_networks, Conv_occupancy_networks, 3D_Unet, NASA} model a probability function of a voxel being occupied by a 3D shape and typically employ a coordinate-based decoder that operates on top of single-view images.
They use the multi-resolution surface extraction method, which is similar in nature to our proposed multi-scale INR.
However, in our case, we share computation between neighboring pixels while they use surface extraction to find regions to refine the predictions on.
In this way, they conduct the full inference for each low-resolution coordinate which is the opposite of what we try to achieve with multi-scale INRs.
DeepSDF \cite{DeepSDF} models a signed distance function instead of the occupancy function and they additionally have an encoder, which transforms an image into a latent code.
IM-NET \cite{IM_NET} proposes to train a generative model on top of these latent codes and conduct experiments not only on ShapeNet objects \cite{ShapeNet}, but on MNIST images as well.
DeepMeta \cite{DeepMeta} models the occupancy function and predicts decoder parameters instead of the latent codes.
A vital branch of research on INRs is concerned about the most efficient way to encode coordinates positions \cite{Transformer, Conv_seq2seq}.
Recent works show that using Fourier features greatly improves INR expressiveness \cite{FourierINR, SIREN}, which we observe in our experiments as well.

\textbf{GANs.}
State-of-the-art results in image generation are held by generative adversarial networks (GANs) \cite{GAN}.
Two key challenges in GAN training are instability and mode collapse, that is why a big part of research in recent years was devoted to finding stronger objective formulation \cite{WGAN, LSGAN, HingeLossGAN}, regularizers \cite{WGAN-GP, R1_reg, Spectral_norm} and architectural designs \cite{ProGAN, StyleGAN, StyleGAN2, MSG_GAN} that would encourage stability, diversity and expressivity of the GAN-based models.

\textbf{Generative models + coordinates.}
There were attempts to combine generative modeling and INR-based representations prior to our work.
IM-NET \cite{IM_NET} trains a generative model on top of latent codes of an occupancy autoencoder.
In our case, instead of feeding a latent code to a coordinate-based decoder, we produce its parameters with a hypernetwork-based generator.
Besides, their image generation experiments were limited to small-scale MNIST-like datasets \cite{MNIST} only.
CoordConv GAN \cite{CoordConv} concatenates coordinates to each representation in the DCGAN model, which endows it with geometric translating behavior during latent space interpolation.
\cite{CoordConv_for_superresolution} use CoordConv layers to perform superresolution.
SBD \cite{Spatial_Broadcast_Decoder} augments a VAE~\cite{VAE} decoder with coordinates information.
SpatialVAE \cite{SpatialVAE} additionally models rotation and translation separately from the latent codes
COCO-GAN \cite{COCO_GAN} generates images by patches given their spatial information and then assembles them into a single image.
\cite{Castle_in_the_sky} uses coordinates-based convolutions for sky replacement and video harmonization.
In their contemporary work, \cite{CIPS} builds an INR-based generator with learnable coordinate embeddings and achieves state-of-the-art results on several large-scale datasets.

\textbf{Hypernetworks.}
Hypernetworks or meta-models are models that generate parameters for other models \cite{Hypernetworks, Hypernetworks_infinite_width}.
Such parametrization provides higher expressivity \cite{Hypernetworks_complexity, Hypernets_modularity} and compression due to weight sharing through a meta-model \cite{Hypernetworks}.
Our factorized multiplicative modulation (FMM) is closely related to the squeeze-and-excitation mechanism~\cite{SENet}.
But in contrast to it, we modulate weights instead of hidden representations, similar to ~\cite{Dynamic_convolutions}, where authors condition kernel weights on an input via attention.
Hypernetworks are known to be unstable to train \cite{Hypernetworks_scale_by_10} and \cite{Hypernetworks_init} proposed a principled initialization scheme to remedy the issue.
In our case, we found it to be unnecessary since our FMM module successfully reduces the influence of hypernetwork initialization on signal propagation inside an INR.
Hypernetworks found many applications in other areas like few-shot learning \cite{Hypernetworks_FSL}, continual learning \cite{Hypernetworks_CL}, architecture search \cite{Hypernetworks_NAS} and others.
In the case of generative modeling, \cite{Hypernetwork_LM} built a character-level language model, and \cite{HyperGAN} proposed a generative hypernetwork to produce parameters of neural classifiers.
Similar to our FMM, \cite{RHH} proposed a low-rank modulation by parametrizing a target model's weight matrix as $\bm W = \bm W_s \odot (\bm{A} \times \bm{B})$, where $\bm W_s$ is a shared component and $\bm A, \bm B$ are rectangular matrices produced by a hypernetwork.

\textbf{Hypernetworks + generative modeling.}
Combining hypernetworks and generative models is not new.
In \cite{HyperGAN} and \cite{Old_HyperGAN}, authors built a GAN model to generate parameters of a neural network that solves a regression or classification task and demonstrate its favorable performance for uncertainty estimation.
HyperVAE \cite{HyperVAE} is designated to encode any target distribution by producing generative model parameters given distribution samples.
HCNAF \cite{Hypernetwork_AF} is a hypernetwork that produces parameters for a conditional autoregressive flow model \cite{Inverse_AF, Masked_AF, Neural_AF}.
\cite{piGAN} trains a generator to produce a 3D object in the INR form and uses mFiLM\cite{Film, FeatureWiseTransformations} to compress the output space.

\textbf{Hypernetworks + INRs.}
There are also works that combine hypernetworks and INRs.
\cite{SRN} represents a 3D scene as an INR, rendered by differentiable ray-marching, and trains a hypernetwork to learn the space of such 3D scenes.
\cite{Hypernetwork_INR} proposes to represent an image dataset using a hypernetwork and perform super-resolution by passing denser coordinate grid into it.
DeepMeta \cite{DeepMeta} builds an encoder that takes a single-view 3D shape image as input and outputs parameters of an INR.
Authors also trained a model to encode a MNIST image into a temporal sequence of digits.


\textbf{Computationally efficient models}.
Among other things, we demonstrate that our INR-based decoder enjoys faster inference speed compared to classical convolutional ones.
Each INR layer can be seen as a $1 \times 1$ convolution \cite{MobileNet, Original_depthwise_sep_convs}, i.e. a convolution with the kernel size of 1.
The core difference is that layer's weights are produced with a hypernetwork $\G$.
Since INR may have a lot of parameters, we factorize them with low-rank factorization~\cite{Lowrank_approx_3}.
There is a vast literature on using low-rank matrix approximations to compress deep models or accelerate them \cite{TTD_NN, Lowrank_approx_1, Lowrank_approx_2, Multi_hashing_model_compression}.
In our case, we found it sufficient just to output a weight matrix as a product of two low-rank matrices $\bm W = \bm{A} \times \bm{B}$ without using any specialized techniques.
SENet \cite{SENet} proposed to apply squeeze-and-excitation mechanism on the hidden representations, and we apply them on the INR weights to make the model more stable to train and faster to converge.

\section{Image Meta Generation}\label{sec:method}

\subsection{Model overview}\label{sec:method:overview}
We adopt the standard GAN training setup and replace a convolutional generator with our INR-based one, illustrated in Figure~\ref{fig:inr-gan-architecture}.
We build upon the StyleGAN2 framework and keep every other component except for the generator untouched, including the discriminator, losses, optimizers, and the hyperparameters.
The details are in Appendix~\apref{ap:implementation-details}.

Our generator $\G$ is hypernetwork-based: it takes latent code $\bm z \sim \mathcal{N}(\bm 0, \bm I)$ as input and generates parameters $\bm\theta$ for an INR model $\F$.
To produce an actual image, we evaluate $\F$ at all the locations of a predefined coordinates grid, which size is determined by the dataset resolution.
For example, for LSUN bedroom $256^2$ we compute pixel values at $256^2 = 65536$ grid locations of $[0,1]^2$ square, that are positioned uniformly.
We use recently proposed Fourier features to embed the $(x,y)$ coordinates \cite{FourierINR, SIREN}.
From the implementation perspective, these Fourier features is just a simple linear layer $\bm u = \sin(\gamma \bm U \bm p)$ with sine (or cosine) activation that maps a raw coordinate vector $\bm p = (x, y)$ into a feature vector $\bm u \in \R^{d_f}$.
Note that our embedding matrix $\bm U$ is not kept fixed and shared but predicted by the generator from $\bm z$.
This gives the model flexibility to select feature frequencies that are the most appropriate for a given image.


\begin{figure}
    \centering
    \begin{subfigure}{\linewidth}
        \centering
        \includegraphics[trim=0 256 0 0,clip,width=\textwidth]{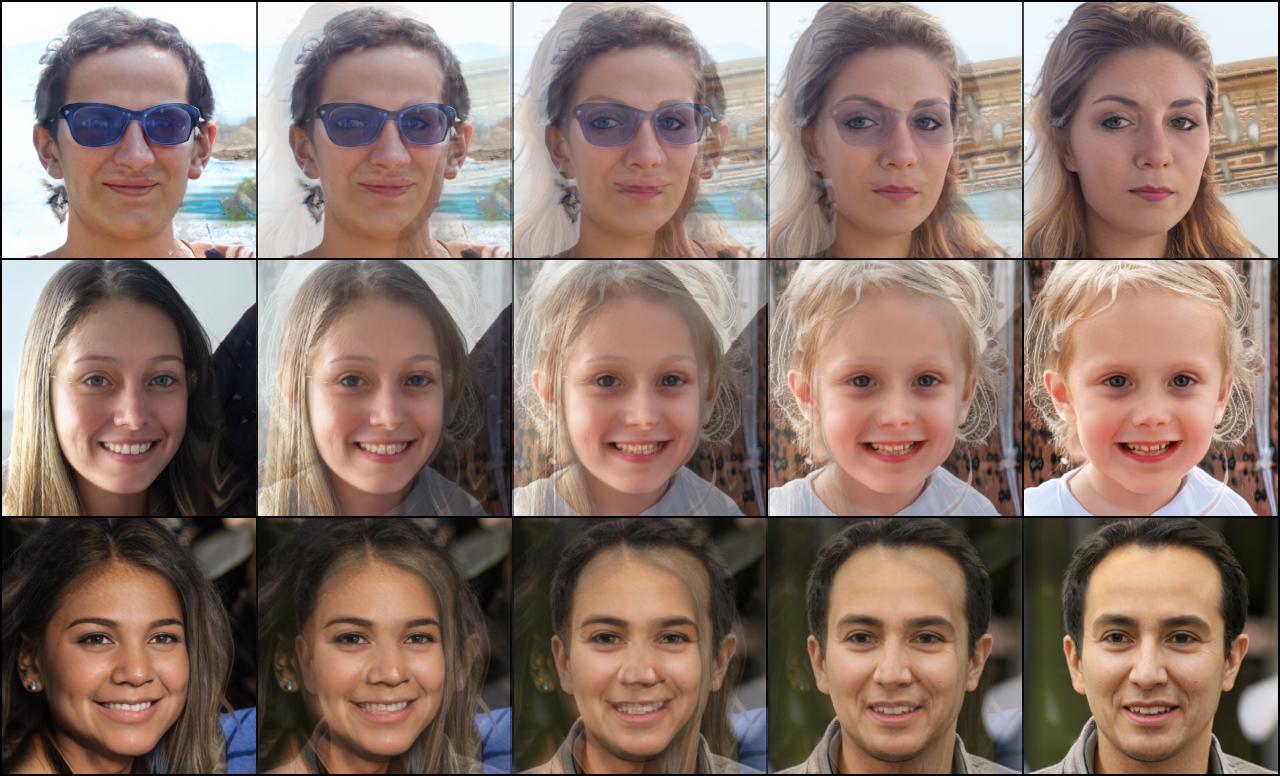}
        \caption{Image interpolation in the pixel-based form.}
        \label{fig:pixel-space-interpolation}
    \end{subfigure}
    \begin{subfigure}{\linewidth}
        \centering
        \includegraphics[trim=0 256 0 0,clip,width=\linewidth]{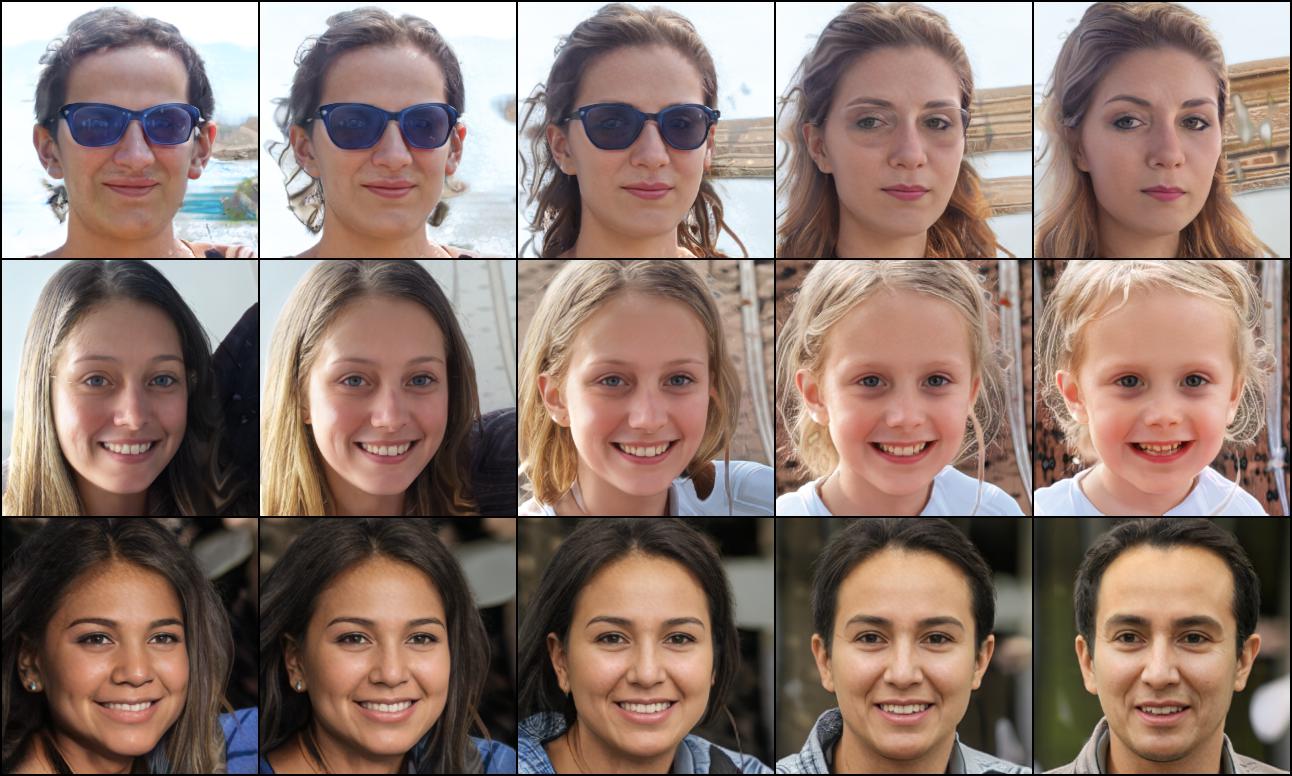}
        \caption{Image interpolation in the INR-based form.}
    \end{subfigure}
    \caption{Images have meaningful interpolation when represented in the INR-based form. To interpolate between $\mathsf{F}_{\bm{\theta_1}}$ and $\mathsf{F}_{\bm{\theta_2}}$, we compute interpolation parameters $\bm{\theta} = \alpha \bm\theta_1 + (1 - \alpha) \bm\theta_2$ and evaluate $\F$ for the provided coordinates grid.}
    \label{fig:inr-space-interpolation}
\end{figure}


\subsection{Factorized Multiplicative Modulation (FMM)}\label{sec:method:fmm}

\begin{figure}
    \centering
    \begin{subfigure}{\linewidth}
    \centering
    \includegraphics[width=0.5\linewidth]{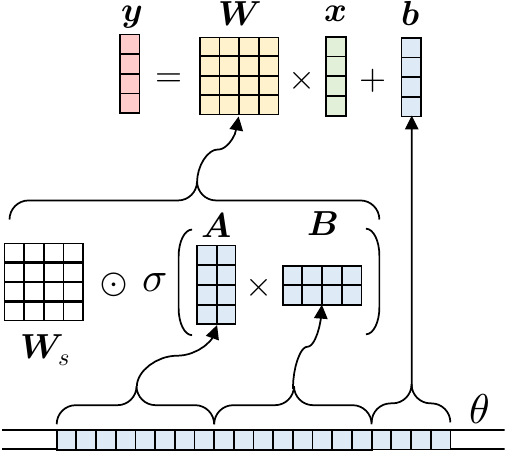}
    \caption{FMM Linear layer.}
    \label{fig:fmm:fmm}
    \end{subfigure}
    
    \vspace{1pt}
    
    \begin{subfigure}{\linewidth}
    \centering
    \includegraphics[width=0.8\linewidth]{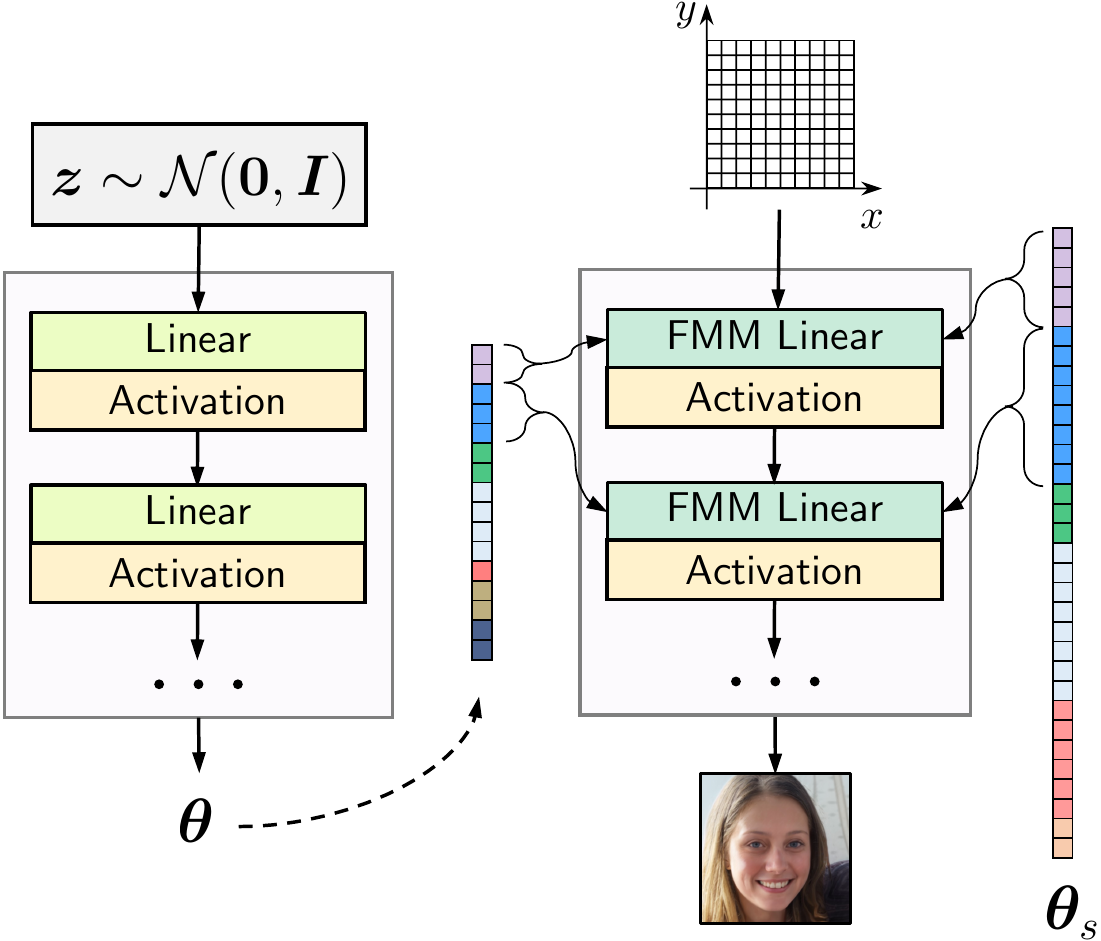}
    \caption{INR-based generator with FMM.}
    \label{fig:fmm:architecture}
    \end{subfigure}
    \caption{(a) FMM linear layer for inputs $\bm x$, shared matrix $\bm W_s$ and output $\bm y$; (b) INR-based generator with the FMM mechanism: its parameters are split into $\bm\theta_s$ (shared) and $\bm\theta$ (predicted by the hypernetwork). This mechanism makes our architecture be similar to StyleGAN2~\cite{StyleGAN2}: the hypernetwork becomes a \textit{mapping network} and the INR $\F$ becomes the \textit{synthesis network} (decoder).}
    \label{fig:fmm}
\end{figure}

A hypernetwork is a model that generates parameters for another model.
In our case, we want to generate INR's parameters given the noise vector $\bm z$.
Imagine, that we need to produce the weights $\bm W^\ell \in \R^{n_\text{in} \times n_\text{out}}, \bm b^\ell \in \R^{n_\text{out}}$ of the $\ell$-th linear layer of $\F$.
A naive implementation would output them directly, but this is extremely inefficient: if the hypernetwork has the hidden dimensionality of size $h$, then its output projection matrix will have the size $h \times (n_\text{in} \cdot n_\text{out} + n_\text{out})$.
Even for small $n_\text{in}, n_\text{out}$ this is prohibitively expensive.

The main problem lies in generating $\bm W^\ell$ since it contains most of the weights, thus factorizing it via low-rank matrix decomposition $\bm W^\ell = \bm A^\ell \times \bm B^\ell$ might seem like a reasonable idea.
However, our preliminary experiments showed that it severely decreases the performance because having low-rank weight matrices is equivalent to having a lot of zero singular values, leading to severe instabilities in GAN training \cite{BigGAN, Spectral_norm}.
That is why we change the parametrization so that the full rank is preserved while the hypernetwork output is still factorized.

Namely, we first define a shared parameters matrix $\bm W_s^\ell \in \R^{n_\text{out}^\ell \times n_\text{in}^\ell}$, which is learnable and shared across all samples.
Our hypernetwork produces two rectangular matrices $\bm A^\ell \in \R^{n_\text{out}^\ell \times r}$ and $\bm B^\ell \in \R^{r \times n_\text{in}^\ell}$, which are multiplied together to obtain a low-rank \textit{modulating} matrix $\bm W_h^\ell = \bm A^\ell \times \bm B^\ell$.
We compute the final weight matrix $\bm W^\ell$ as $\bm W^\ell = \bm W_s^\ell \odot \sigma(\bm W_h^\ell)$ where $\sigma$ is sigmoid function.
Bias vector $\bm b^\ell$ is produced directly since it is small. 
In all the experiments, we set $r=10$ for all the layers of $\F$ except the first and the last ones which are small enough to be generated directly.
We illustrate FMM in Figure~\ref{fig:fmm}.

The above modulation is very similar to the one proposed in \cite{RHH} with the difference that \cite{RHH} does not use the sigmoid activation.
However, as our experiments in Section~\ref{sec:experiments} demonstrate, using the activation is crucial since it bounds the activations and makes the training more stable.
Note that the FMM-based INR-GAN becomes very close architecturally to StyleGAN2 \cite{StyleGAN2}, which uses a mapping network to predict the style vector used to modulate the convolutional weights of its decoder via multiplication.

\subsection{Multi-scale INRs}\label{sec:method:multi-scale}

\begin{figure}
\centering
\includegraphics[width=0.8\linewidth]{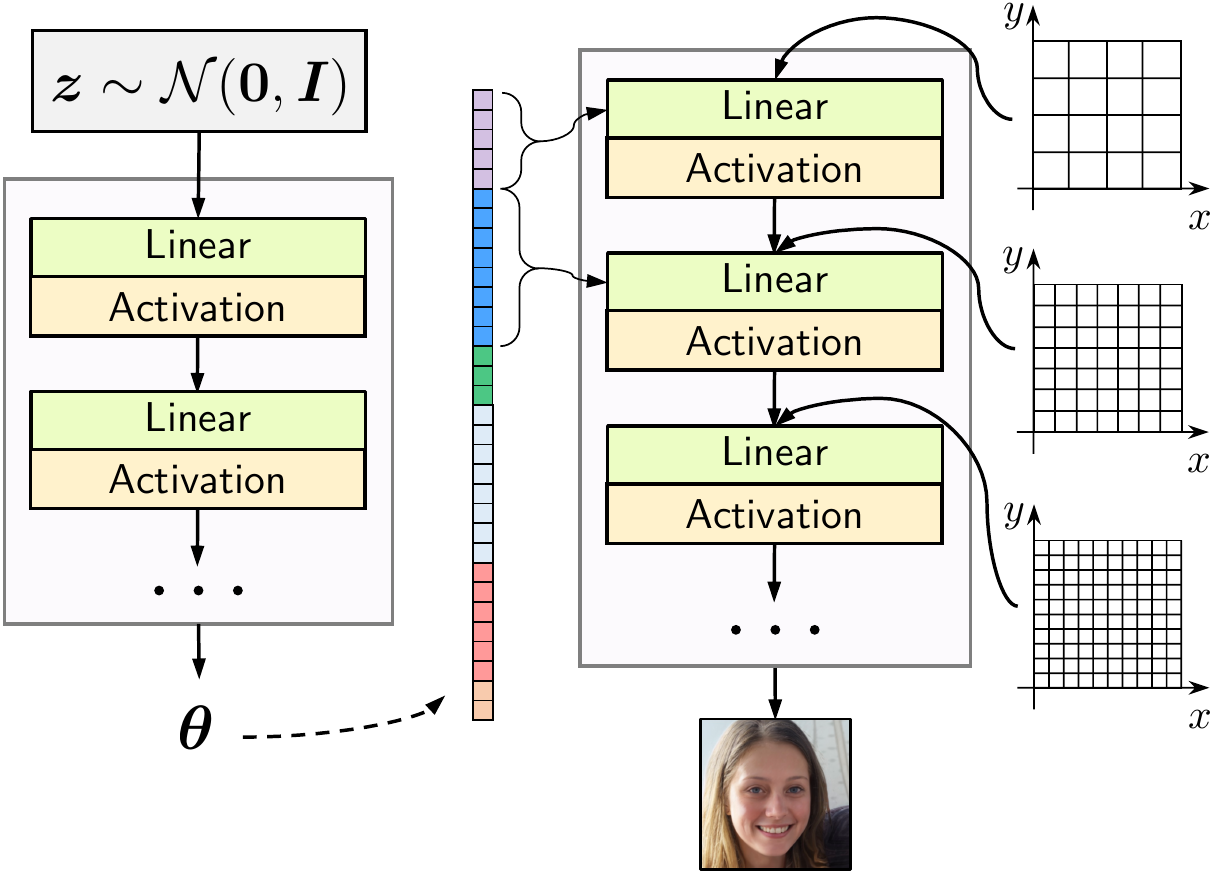}
\caption{\textbf{Multi-Scale INR-based GAN (without FMM)}. Each block operates on a different resolution, determined by the granularity of an input grid. We increase the granularity with depth: this allows to share computation between neighbouring pixels \textit{and} condition them on a common context. We depict the multi-scale mechanism without FMM not to clutter the illustration. In practice, we use both FMM and the multi-scale architecture for our INR-GAN.}
\label{fig:multi-scale-inr}
\end{figure}

Scaling a traditional INR-based decoder to large image resolutions is too expensive: to produce a $1024^2$ image, we have to input ${\approx}10^6$ coordinates into $\F$.
Such a huge batch size makes it impossible to use large hidden layers' sizes due to excessive computation and memory consumption.
To circumvent the issue, we propose a \textit{multi-scale INR} architecture: we split $\F$ into $K$ blocks, where each block operates on its own resolution and only the final block operates on the target one.
Earlier blocks compute low-res features that are then replicated and passed to the next level.
This process is illustrated on ~\figref{fig:multi-scale-inr} and is equivalent to using different grid sizes depending on the resolution and using the nearest neighbor interpolation for the hidden representations.
For the multi-scale INRs, we use more neurons for lower resolutions and fewer ones for the more expensive high-resolution blocks.
The use of multi-scale INRs makes our architecture very similar to classical convolutional decoders, which grow the resolution progressively with depth \cite{distill_deconvs}.
In our case, one of the additional benefits of having the multi-scale architecture is that neighboring pixels get conditioned on the common context computed at a previous resolution.

We start with the resolution of $64^2$ for the first block and increase it by 2 for each next one until we reach the target resolution.
Each block contains 2-4 layers, and Fourier coordinates features are concatenated to a hidden representation at the beginning of each block.
Replicating low-resolution features for each high-dimensional pixel is equivalent to upsampling the inner grid with nearest neighbors interpolation, and this is the way we implement it in practice.
Further implementation details can be found in the supplementary material or the attached source code.

\section{Experiments}\label{sec:experiments}

\subsection{Standard GAN training}


\begin{table*}
\caption{We start with the INR-based decoder conditioned on the latent code \cite{IM_NET, SpatialVAE} and progressively improve it.
O/M denotes ``out-of-memory'' error: we couldn't train the model even for a batch size of 1 on a 32GB NVidia V100 GPU.}
\label{table:main-results}
\centering
\resizebox{1.0\linewidth}{!}{
\begin{tabular}{|l|lll|lll|lll|}
\hline
\multirow{2}{*}{Decoder type} & \multicolumn{3}{c|}{LSUN $128^2$} & \multicolumn{3}{c|}{LSUN $256^2$} & \multicolumn{3}{c|}{FFHQ $1024^2$} \\
& GMACs & \#params & FID & GMACs & \#params & FID & GMACs & \#params & FID \\
\hline
Latent-code conditioned INR decoder \cite{GAN_for_implicit_shapes, SpatialVAE} & 30.09 & 7.1M & 229.9 & 120.33 & 7.1M & 253.3 & 1925.22 & 7.1M & O/M \\
+~Hypernetwork-based decoder \cite{DeepMeta} & 23.54 & 2055.5M & 28.83 & 88.01 & 2055.5M & O/M & 1377.52 & 2055.5M & O/M \\
+~Fourier embeddings from \cite{SIREN, FourierINR} \ours & 28.02 & 2312.02M & 23.07 & 105.34 & 2312.02M & O/M & 1651.52 & 2312.02M & O/M \\
+~Factorized Multiplicative Modulation \ours & 25.87 & 108.2M & 11.51 & 103.19 & 108.2M & 15.68 & 1649.37 & 108.2M & O/M \\
+~Multi-Scale INR \ours & 21.58 & 107.03M & 5.69 & 38.76 & 107.03M & 6.27 & 47.35 & 117.3M & 16.32 \\
\hline
StyleGAN2 generator \cite{StyleGAN} & - & - & - & 84.36 & 30.03M & 2.65 & 143.18 & 30.37M & 4.41 \\
\hline
\end{tabular}
}
\end{table*}

\textbf{Datasets}.
We conduct experiments on four datasets: LSUN Bedrooms $128^2$, LSUN Bedrooms $256^2$, LSUN Churches $256^2$ and FFHQ $1024^2$.
LSUN Bedrooms and LSUN Churches consist on 3M and 125k images of in-the-wild bedrooms and churches \cite{LSUN}, respectively.
FFHQ is a high-resolution dataset of 70k human faces \cite{StyleGAN}.
For all the datasets, we apply random horizontal flip for data augmentation.

\textbf{Evaluation metrics}.
We evaluate the model using Frechet Inception Distance (FID) \cite{TTUR_FID} metric using 50k images to compute the statistics.
We also compute the number of parameters for a given model and the amount of multiply-accumulate operations (MACs) --- a standard measure for accessing the model's computational efficiency \cite{MobileNet, SlowFast}.

\textbf{Models and training details}.
All our models have equivalent training settings and differ only in the generator architecture.
We use two existing architectures as our baseline:
\begin{enumerate}
    \item A non-hypernetwork-based generator, which has shared parameters for all INRs and each INR is conditioned on latent code $\bm w = \G(\bm z)$ \cite{IM_NET, SpatialVAE, DeepSDF}. I.e., instead of producing parameters for $\F$, we pass $\bm w$ into it as an additional input $\bm v = \F(x, y, \bm w)$. 
    \item A hypernetwork-based generator, which produces parameters $\bm\theta$ for $\F$ but does not use any factorization techniques to reduce the output matrix size \cite{DeepMeta}.
\end{enumerate}
We build upon the above baselines by incorporating Fourier positional embeddings of the coordinates \cite{FourierINR, SIREN}, incorporating our FMM layer and incorporating our multi-scale INR architecture.
In all the experiments, $\G$ is a 4-layer MLP with residual connections that takes $\bm z \in \R^{512}$ as input and produces INR parameters $\bm\theta$ as output.
For the first baseline, it produces the transformed latent codes instead.
For hypernetwork-based models, we additionally apply a learned linear layer to $\bm w$ to obtained INR parameters $\bm \theta$.
In all the experiments, a ResNet-based discriminator from StyleGAN2~\cite{StyleGAN2} is used.
However, since beating the scores is not the goal of the paper, we used its ``small'' version (corresponding to config-e).
$R_1$-regularization \cite{R1_reg} with the penalty weight of 10 is used.
All the models are trained for 800k iterations on 4 NVidia V100 32GB GPUs.

\textbf{Results}.
The results are reported in Table~\ref{table:main-results}.
As we can see, the latent code conditioned baseline cannot fit training data at all since it lacks expressivity and cannot capture the whole image representation in the latent code.
Its hypernetworks-based counterpart achieves much higher performance but is still not competitive.
An attempt to fit the baselines on FFHQ $1024^2$ resulted in out-of-memory errors even for a batch size of 1 for a 32GB GPU.
The non-factorized hypernetwork-based decoder \cite{DeepMeta} was too expensive even for $256^2$ resolution.

Our INR-based decoder with FMM layer and multi-scale INR architecture achieves competitive FID scores and greatly reduces the gap between continuous and pixel-based image generation.
It is still inferior to StyleGAN2 \cite{StyleGAN2} in terms of performance, but is three times more efficient.
One should also note that we didn't use any of the numerous training tricks employed by StyleGAN2 and that our discriminator architecture is smaller (it corresponds to config-e \cite{StyleGAN2}).
Besides, as we show in Section \ref{sec:experiments:properties}, our model naturally gives rise to a lot of additional properties that convolutional decoders lack.
Our model also has three times more parameters than its convolution-based counterpart.
The reason for it is the huge size of the output projection matrix of $\G$, which has the dimensionality of $\approx 10^3 \times 10^5$ occupying 90-99\% of parameters of the entire model.
Compressing this matrix is an ongoing hypernetworks research topic \cite{Hypernetworks, Multi_hashing_model_compression} and we leave this for future work.
We emphasize that despite its enormous size, the output layer's inference time is almost negligible due to highly optimized matrix-matrix multiplications on modern GPUs.

\begin{figure}
    \centering
    \includegraphics[width=0.8\linewidth]{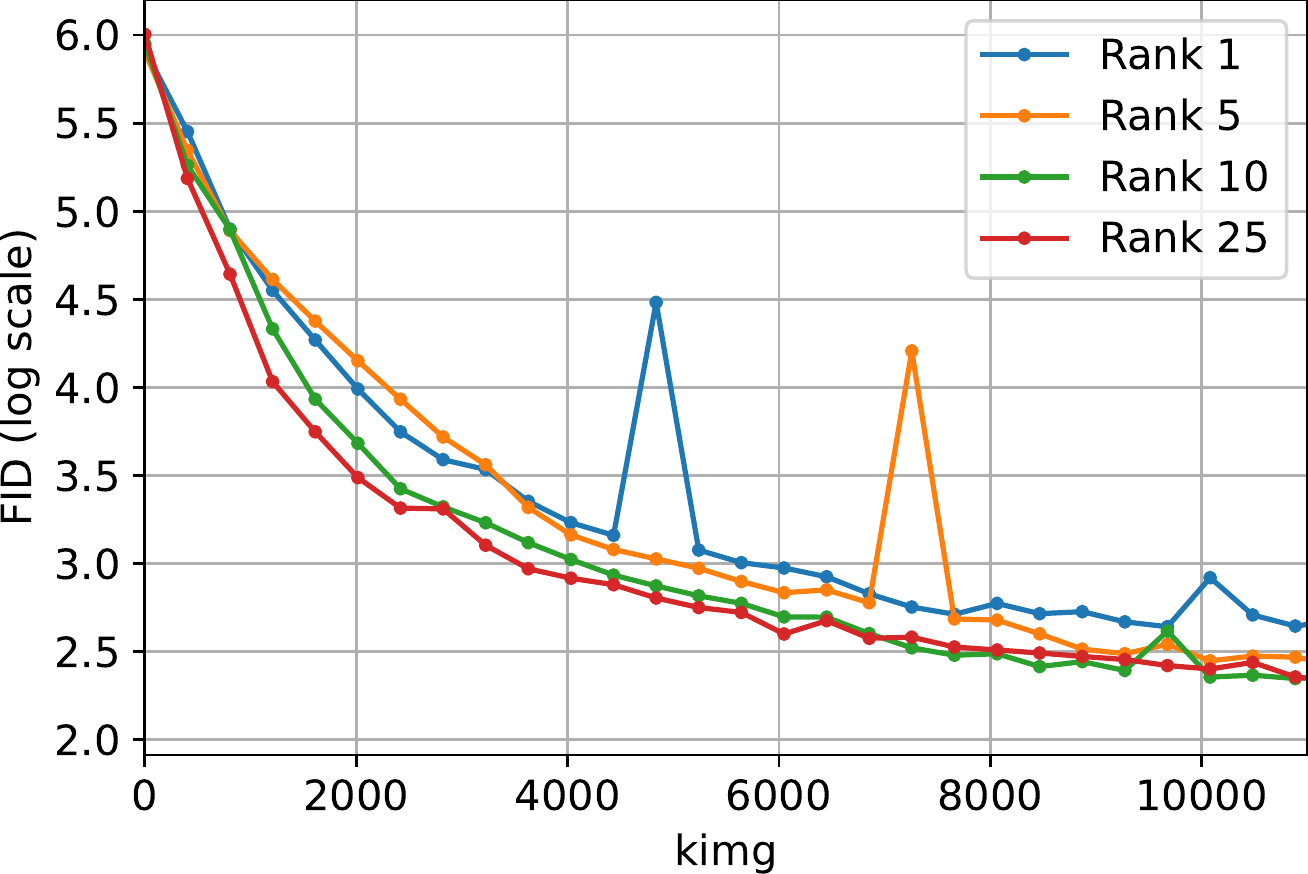}
    \caption{FID scores (in log scale) on LSUN Bedroom $256^2$ for different rank values of the proposed FMM modulation. Increasing the rank beyond 1 gives a clear advantage, but it diminishes for the subsequent rank increase. ``kimg'' denotes the number of images seen by $\D$.}
    \label{fig:fmm-ranks-ablation}
\end{figure}

\textbf{Ablating FMM rank}.
In all the experiments, we use an FMM rank of 10.
We ablate over its importance for LSUN bedroom $256^2$ and report the resulting convergence plots on ~\figref{fig:fmm-ranks-ablation}.
As one can see, the model benefits from the increased rank, but the advantage is diminishing for further rank increase.
These results show that vanilla or "full-rank" hypernetworks are heavily overparameterized and there is no need to predict each parameter separately. On the other that is the evidence that there is much potential for "going further" than squeeze-and-excitation \cite{SENet} and AdaIN \cite{AdaIN, StyleGAN} approaches, i.e. predicting more than a single modulating value for each neuron.

\begin{table}
\caption{FID scores for additional ablations of our multi-scale INR-based GAN (INR-GAN).
Removing $\sigma(x)$ from FMM \textit{worsens} the scores.
Incorporating StyleGAN2's architecture and bilinear upsampling allows the INR-based generator to rival its convolution-based counterpart.
Img/sec of $\G$ is measured on 1 NVidia V100 32GB.}
\label{table:improved-results}
\centering
\resizebox{1.0\linewidth}{!}{
\begin{tabular}{|l|cc|c|}
\hline
Decoder type & Churches $\downarrow$ & Bedrooms $\downarrow$ & img/sec $\uparrow$ \\
\hline
INR-GAN w/o FMM activation & 10.35 & 11.73 & \textbf{267.3} \\
INR-GAN & 7.12 & 6.27 & 267.1 \\
~+ StyleGAN2 architecture & 5.09 & 4.96 & 265.1 \\
~+ bilinear upsampling & \textbf{3.12} & 3.41 & 203.4 \\
\hline
StyleGAN2 & 3.86 & \textbf{2.65} & 85.5 \\
\hline
\end{tabular}
}
\end{table}

\textbf{Additional ablations}.
All the above experiments were conducted on top of a vanilla INR-based decoder which uses nearest neighbour interpolation (to make neighboring pixels be computed independently as the vanilla INR does).
It also lacks important StyleGAN2's techniques which improve the performance, like equalized learning rate \cite{ProGAN}, style mixing, pixel normalization \cite{ProGAN} and noise injection \cite{StyleGAN}.
In all our experiments, we also used a small-size version of StyleGAN2's discriminator to make the training run faster.
Incorporating the above design choices and giving up the nearest neighbour upsampling for the bilinear one (which might be reasonable for some applications) significantly improves our generator's performance.
We conduct those experiments on LSUN Churches $256^2$ and LSUN Bedrooms $256^2$ and report the results in Table~\ref{table:improved-results}.
For the INR-based decoder with bilinear upsampling, we even surpass StyleGAN2's performance on Churches.

Another critical question is how important the activation in FMM is.
As discussed in Section~\ref{sec:method:fmm}, it allows to stabilize training by bounding the magnitudes of the weights.
We ablate its influence by replacing $\sigma$ with the identity mapping and report the resulted FID in Table~\ref{table:improved-results}.
The drop in performance demonstrates that it carries a crucial influence on the image quality.

\textbf{Performance on multi-class datasets}.
To test if the proposed architectural techniques improve the performance on diverse multi-class datasets, we also conduct experiments on LSUN-10 $256^2$ and MiniImageNet $128^2$.
We provide the details of these experiments in Appendix~\apref{ap:diverse-datasets}.


\subsection{Exploring the properties}\label{sec:experiments:properties}

\textbf{Extrapolating outside of image boundaries}.
At test time, we sample pixels beyond the coordinates grid that the model was trained on and present the results on ~\figref{fig:oob-generalization}.
It shows that an INR-based decoder is capable of generating meaningful content outside of the grid boundaries, which is equivalent to a zooming-out operation.
It is a surprising quality indicating that the coordinates features produced by the generator are exploited by an INR in a generalizable manner instead of just being simply memorized.

\textbf{Meaningful interpolation}.
Image space interpolation is known for its poor behavior \cite{deep_learning_book}.
However, when images are represented in the INR-based form and not the pixel-based one, the interpolation becomes reasonable.
We illustrate the difference on ~\figref{fig:inr-space-interpolation}.

\textbf{Keypoints prediction}.
Direct access to coordinates provides more fine-grained control over the geometrical properties of images during the generation process.
Thus one can expect the INR-based decoder to better embed the geometric structure of an image into the latent space \cite{CoordConv}.
To test this hypothesis, we perform the following experiment.
We take 10k samples from our model trained on FFHQ $256^2$ and extract face keypoints with a pre-trained model \cite{SuperFAN}.
After that, we fit a simple linear regression model to predict these keypoints coordinates given the corresponding latent code.
We compute this model's test loss on the real-world images from FFHQ $256^2$ and coin this metric Keypoints Prediction Loss (KPL).
The above procedure is formally described in Algorithm~\apref{alg:kpl} in Appendix~\apref{ap:experiments-details}.
The corresponding latent codes for the real-world images are obtained by projecting an image into the corresponding latent space through gradient descent optimization.
We use the standard protocol from \cite{StyleGAN2} to do this for both models.
KPL is computed for both $\mathcal{Z}$-space and $\mathcal{W}$-space.
For our model, this corresponds to $\G$ input noise vectors $\bm z$ and penultimate hidden representations (i.e. hidden representations before the output projection into INR parameter space).
For StyleGAN2, this corresponds to the mapping network's input and output space.
We also compute KPL on top of random vectors to check that a better performance in keypoints prediction is not due to their reduced variability.
The results of this experiment are provided in Table~\ref{table:keypoints-prediction}.

\begin{table}
\centering
\caption{Keypoints Prediction Loss (KPL) metric for our INR-based decoder and StyleGAN2 generator computed on FFHQ $256 \times 256$ dataset for different latent spaces. The qualitative difference is illustrated on ~\figref{fig:keypoints-prediction}.}
\label{table:keypoints-prediction}
\resizebox{1.0\linewidth}{!}{
\begin{tabular}{| c | c | c | c |} 
 \hline
 Generator & KPL $(\mathcal{Z})$ & KPL $(\mathcal{W})$ & KPL (random) \\ [0.5ex] 
 \hline
 StyleGAN2 & $2.1 \cdot 10^{-4} $ & $7.6 \cdot 10^{-5}$ & $4.6 \cdot 10^{-4}$ \\ 
 INR-GAN & $6.0 \cdot 10^{-5}$ & $5.2 \cdot 10^{-5}$ & $4.7 \cdot 10^{-4}$ \\
 \hline
\end{tabular}
}
\end{table}

\begin{figure}[t!]
    \centering
    \includegraphics[width=0.7\linewidth]{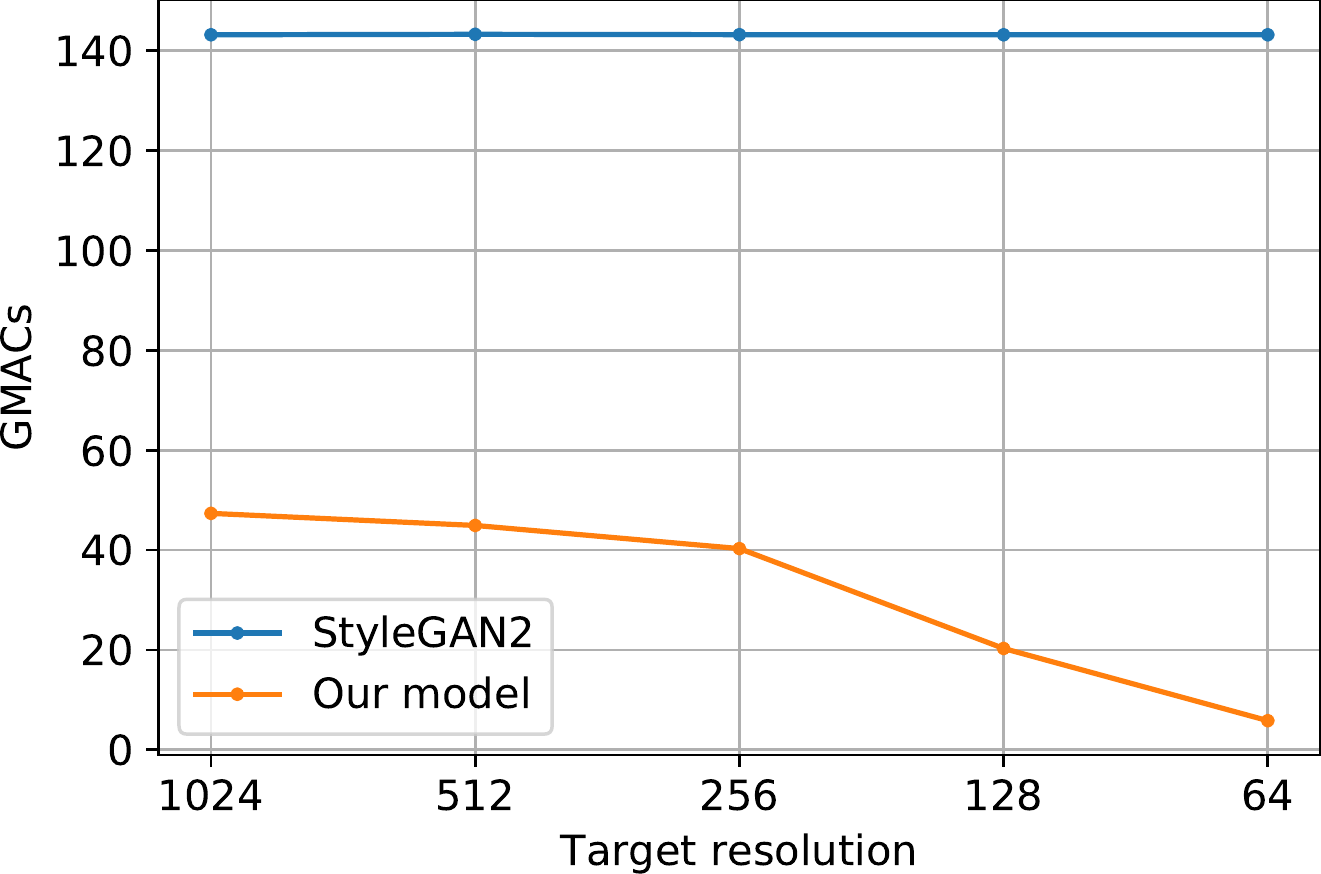}
    \caption{\textbf{Accelerated low-resolution image generation}. We measure a decoder's efficiency in terms of \#MACs on generating an image of lower resolution compared to what it has been trained on. Since INR can do this by evaluating on a sparser grid, this allows it to save a lot of computation. Traditional convolutional decoders require performing a full inference first and then downsampling the produced image.}
    \label{fig:fast-lowres-generation}
\end{figure}

\textbf{Accelerated inference of lower-resolution images}.
Our INR-GAN has a natural capability of generating a low-resolution sample faster because it can be directly evaluated at a low-resolution coordinates grid without performing the full generation.
However, traditional convolutional decoders lack this property: to produce a lower-resolution image, one would need to perform the full inference and then downsample the resulted image with standard interpolation techniques.
To state the claim rigorously, we compute an amount of multiply-accumulate (MAC) operations for our INR-based generator and StyleGAN2 generator trained on FFHQ $1024^2$ for different lower-resolution image sizes.
To produce the low-resolution image with StyleGAN2, we first produce the full-resolution image and them downsample it with nearest neighbour interpolation.
For our model, we just evaluate it on a grid of the given resolution.
The results are reported on ~\figref{fig:fast-lowres-generation}.
To the best of our knowledge, our work is the first one that explores the accelerated generation of lower-resolution images --- an analog of early-exit strategies \cite{BranchyNet} for classifier models for image generation task.

\textbf{Out-of-the-box superresolution}.
Our INR-based decoder is able to produce images of higher resolution than it was trained on.
For this, we evaluate our model on a denser coordinates grid.
To measure this quantitatively, we propose UpsampledFID: a variant of FID score where fake data statistics are computed based on upsampled low-resolution images.
We compare our UpsampledFID score to three standard upsampling techniques: nearest neighbor, bilinear, and bicubic interpolation.
\figref{fig:superresolution} demonstrates that INR-upsampling improves the score by up to 50\%.


\section{Additional potential of INR-based decoders}
In Section ~\ref{sec:experiments:properties}, we explored several exciting properties of INR-based decoders and tested them in practice.
Here, we highlight their additional potential and leave its exploration for future work.

\textbf{An ability to backpropagate through pixels positions}.
Since coordinate positions $(x,y)$ are transformed via well-differentiable operations, it gives a possibility to back-propagate through coordinate positions.
This opens a large range of possible work like using spatial transformer layers \cite{STN} in the generator and not only discriminator.
Or producing a coordinates grid with the discriminator for zooming-in into specific parts of an image.

\textbf{Faster inference speed}.
Since INRs do not use local context information during inference, it does not spend computation on aggregating it as convolutions do.
This is equivalent to using convolutions with a kernel size of 1 and thus works much faster \cite{MobileNet}.
Table~\ref{table:main-results} demonstrates that our model uses three times fewer MACs for its inference process, which is due to ignoring context information.

\textbf{Parallel pixel computation}
Non-autoregressive models like Parallel WaveNet \cite{ParallelWaveNet, Non_autoregressive_MT} are valued by their ability to generate an arbitrarily long sequence in parallel, so their inference speed decreases linearly with more compute being added.
INR-based decoders also have this property due to their independent pixel generation nature.
Convolutional decoders, in contrast, are forced to use local context during the inference process, which limits their parallel inference.

\textbf{Universal decoder architecture}
One can employ the same decoder architecture for training a decoder model in any domain: 2D images, 3D shapes, video, audio, etc.
They would only differ at what coordinates are being passed as input to the corresponding INR.
For 2D images, these would be $(x, y) \R^2$ coordinates, for 2D video, this would be $(x, y, t) \R^2 \times \R_+$ with the additional timestep $t \in \R_+$ input, for 3D shapes --- $(x, y, z) \in \R^3$ coordinates, etc.
That has already been partially explored in \cite{DeepMeta, SIREN}.


\textbf{Biological plausbility}.
While convolutional \textit{encoders} have a very intimate connection to how a human eye works \cite{lecun_convs}, convolutional \textit{decoders} do not have much resemblance to any human brain mechanism.
In contrast, \cite{DL_cortex} argue that hypernetworks have a very close relation to how a prefrontal cortex influences other brain parts.
It does so by modulating the activity in several different areas at once, precisely how our hypernetwork-based generator applies modulation to different INR layers.

\section{Limitations}
The core limitation of INR-based decoders comes from their strengths: not using spatially common context between neighboring pixels.
Multi-Scale INR architecture partially alleviates this issue by grounding them on the same low-resolution representation, but as our experiment with adding bilinear interpolation demonstrates (see Table~\ref{table:improved-results}), it is not enough.
Also, we noticed that the INR-based decoders might become too sensitive to high-frequency coordinates features.
This may produce ``texture'' artifacts: a generated image having a random transparent texture spanned on it which becomes more noticeable for higher resolutions.

\section{Conclusion}
This paper explored the adversarial generation of continuous images represented in the implicit neural representations (INRs) form.
We proposed two principled architectural techniques: factorized multiplicative modulation and multi-scale INRs, which allowed us to obtain solid state-of-the-art results in continuous image generation.
We explored several attractive properties of INR-based decoders and discussed their future potential and limitations.

{\small
\bibliographystyle{ieee_fullname}
\bibliography{egbib}}

\clearpage
\newpage

\appendix
\onecolumn
\section{Additional implementation details}\label{ap:implementation-details}
We build on top of the StyleGAN2 framework \cite{StyleGAN2} and change only its generator.
All other settings, including the discriminator architecture $\D$, optimizers, losses, training settings and other hyperparameters are kept untouched.
This means that we use non-saturating logistic loss for training \cite{GAN} and use Adam optimizers with the parameters $\beta_1 = 0.0, \beta_2 = 0.98$ and $\epsilon = 1e-8$.
Our $\D$ is a small version of StyleGAN2 discriminator (i.e. config-e from the paper \cite{StyleGAN2}), regularized with zero-centered $R_1$ gradient penalty \cite{R1_reg} with weight $\gamma = 10$.
We apply the regularization on each iteration instead of using the lazy setup, as done by StyleGAN2\cite{StyleGAN2} who applies it only each 16-th iteration.
We use the learning rate of 0.00001 for $\G$, 0.0005 for the shared parameters of an INR (which is a part of $\G$) and 0.003 for $\D$.
We also employ skip-connections for coordinates inside each multi-scale INR block \cite{MSG_GAN}.
We apply them by concatenating the coordinates to inner representations.

For the main experiments, we didn't employ any ProGAN \cite{ProGAN}, StyleGAN \cite{StyleGAN} or StyleGAN2 \cite{StyleGAN2} training tricks, like path regularization, progressive growing, equalized learning rate, noise injection, pixel normalization, style mixing, etc.
For our additional ablations, we reimplemented our INR-based decoder on top of the StyleGAN2's generator and employed style mixing, equalized learning rate and pixel normalization for it.
In terms of implementation, it was equivalent to replacing StyleGAN2's weight modulation-demodulation with our FMM mechanism, reducing kernel size from 3 to 1, concatenating coordinates information at each block and replacing its upfirdn2d upsampling with the nearest neighbour one.
We found that for the nearest neighbour upsampling, the model learns to ignore spatial noise injection (by setting noise strengths to zero), because pixels cannot communicate the noise information between each other, making it meaningless and harmful to the generation process.

For our $256 \times 256$ experiments (for both LSUN and FFHQ), we use 2 multi-scale INR blocks of resolutions 128 and 256.
Each block contains 4 layers of 512 dimensions each.
For FFHQ $1024 \times 1024$, we used 4 multi-scale INR blocks of resolutions 128, 256, 512 and 1024.
First 2 blocks had 3 layers of dimensionality of 512, 512 resolution had 2 layers of resolution 128, the final block had 2 layers of dimensionality of 32.

Our $\G$ architecture is the same for all the experiments and consists on 3 non-linear layers with the hidden dimension of 1024 and residual connections.
Our noise vector $\bm z$ has the dimensionality of 512.

For additional implementation details, we refer a reader to the accompanying source code.

In all the experiments, we compute FID scores based on 50k images using the tensorflow script from BigGAN pytorch repo\footnote{\url{https://github.com/ajbrock/BigGAN-PyTorch/blob/master/inception_tf13.py}}.

\section{Experiments details}\label{ap:experiments-details}

\subsection{Zooming out}
\begin{figure*}
    \centering
    \includegraphics[width=0.9\textwidth]{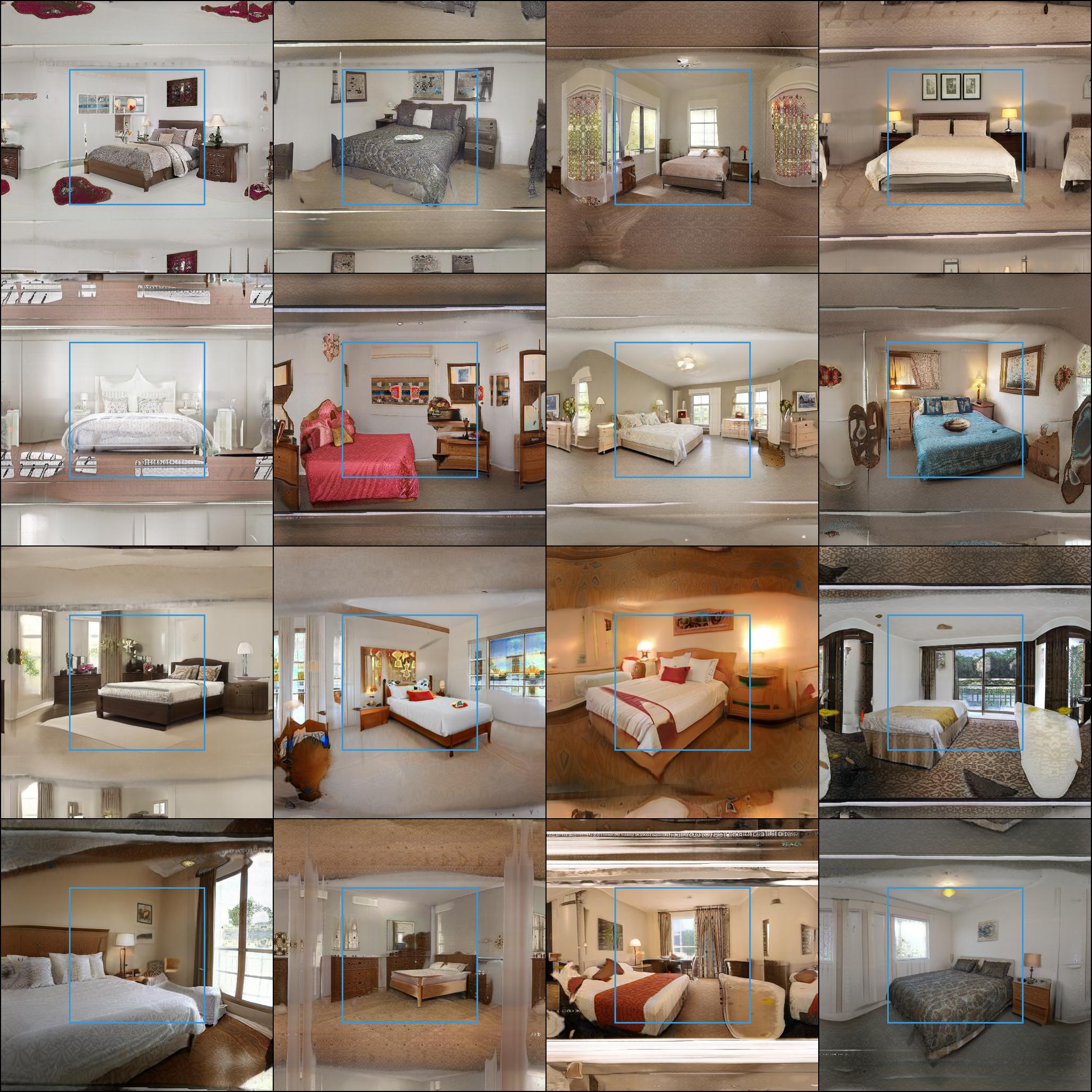}
    \caption{After training our INR-based GAN on LSUN $256 \times 256$ dataset, we feed larger coordinates grid $[-0.5, 1.5]^2$ into it. This is larger than on Figure~\ref{fig:oob-generalization} to demonstrate the extent to which the model extrapolates. As one can see, extending the grid outside of $[-0.4, 1.4]^2$ makes the quality to decrease rapidly.}
    \label{fig:zoom:extra}
\end{figure*}

On ~\figref{fig:zoom:extra} we present additional examples of our zooming operation, but evaluating the model on $[-1.5, 1.5]^2$ grid instead of $[-0.3, 1.3]^2$ like we did for ~\figref{fig:oob-generalization} in the main body.
This is done to demonstrate the extent to which the model can extrapolate.

\subsection{Keypoints prediction}
As being said, we train a linear model to predict the keypoints from the latent codes.
To train such a model, we first generate $n=10^4$ latent codes $\bm w_1, ..., \bm w_n$ for each model, then we decode them into images $\bm x_1, ..., \bm x_n$.
After that, we predict keypoints vector $\bm y_i$ for each image with Super-FAN model \cite{SuperFAN}.
Then we fit a linear regression model to predict $\bm y_i$ from $\bm w_i$.

Measuring quality based on the synthesized images may be unfair since the variability in keypoints of each model can be different: imagine a generator that always produces a face with the same keypoints.
This is why we compute the test quality by embedding real FFHQ images into each model.
But to additionally demonstrate that the variability is equal for the both models, we fit a linear regression model on \textit{randomly permuted} latent codes and check its score: if the prediction accuracy is high, than the variability in keypoints is low and the prediction task is much easier.
We call this metric KPL (random) and depict the corresponding values on Table \ref{table:keypoints-prediction}.
It clearly demonstrates that the both models have equal variability in terms of keypoints.

\begin{figure*}
    \centering
    \begin{subfigure}[b]{0.49\linewidth}
    \centering
    \includegraphics[width=0.3\textwidth]{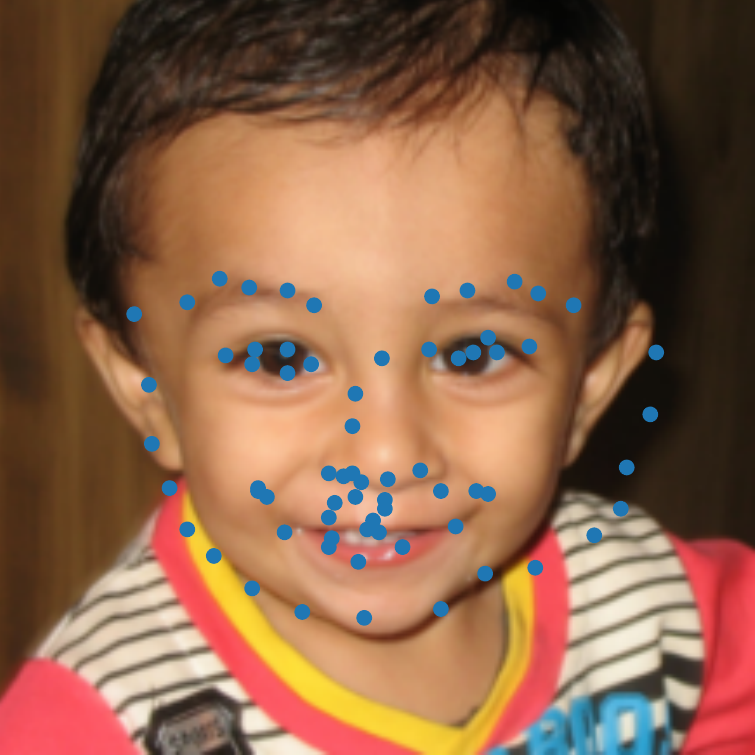}
    \includegraphics[width=0.3\textwidth]{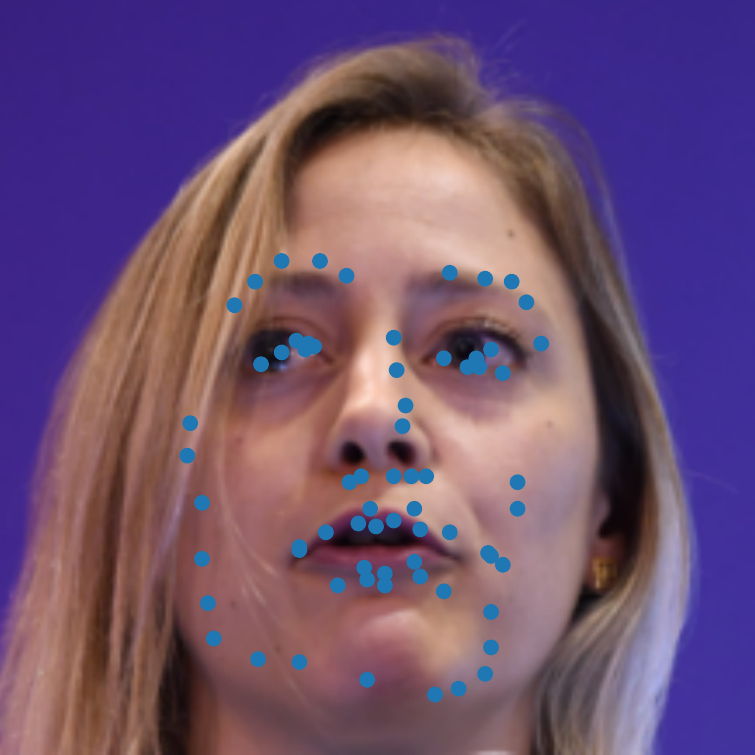}
    \includegraphics[width=0.3\textwidth]{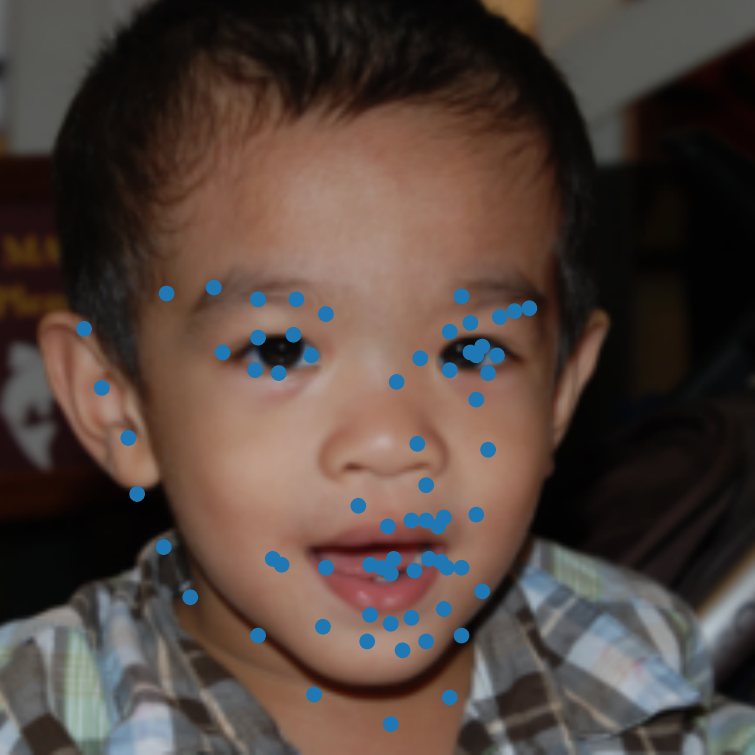}
    \includegraphics[width=0.3\textwidth]{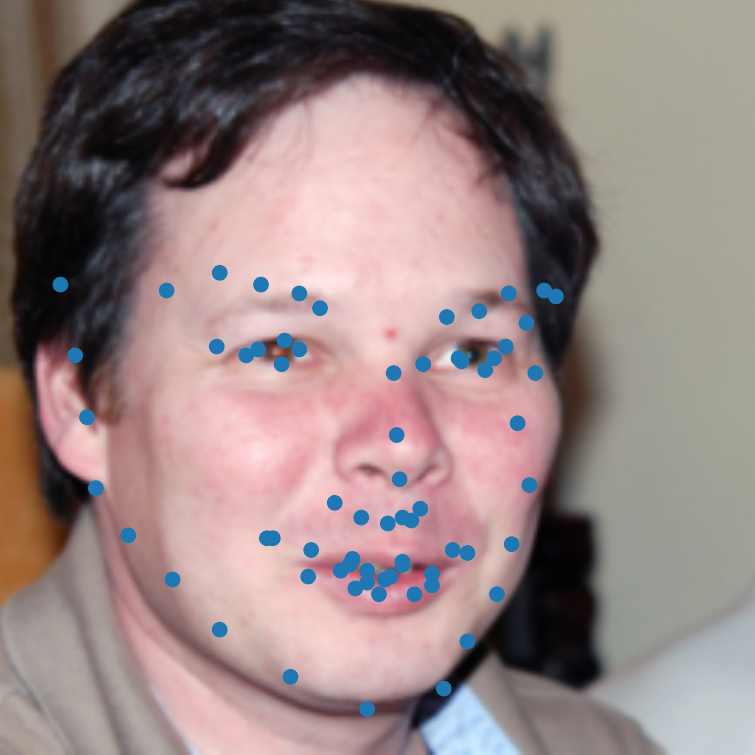}
    \includegraphics[width=0.3\textwidth]{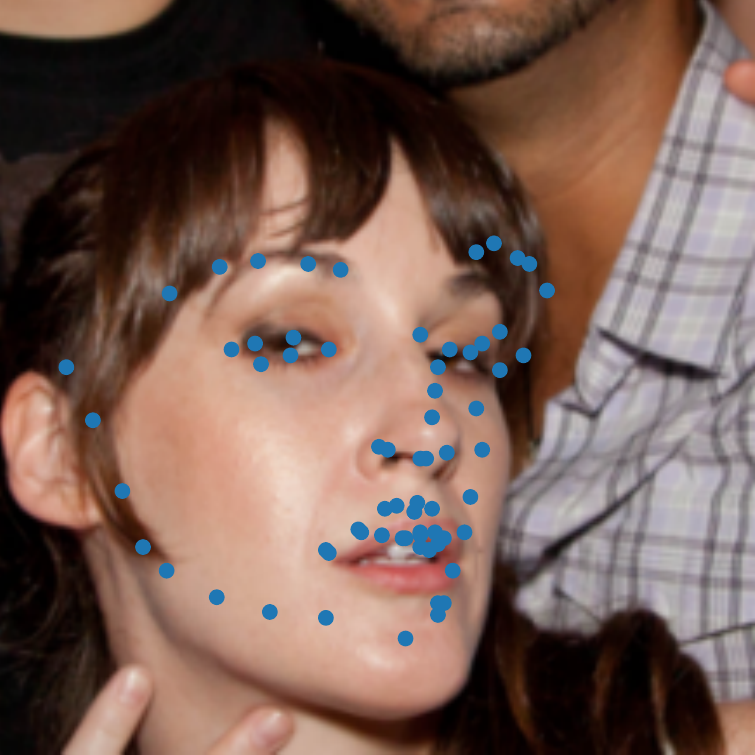}
    \includegraphics[width=0.3\textwidth]{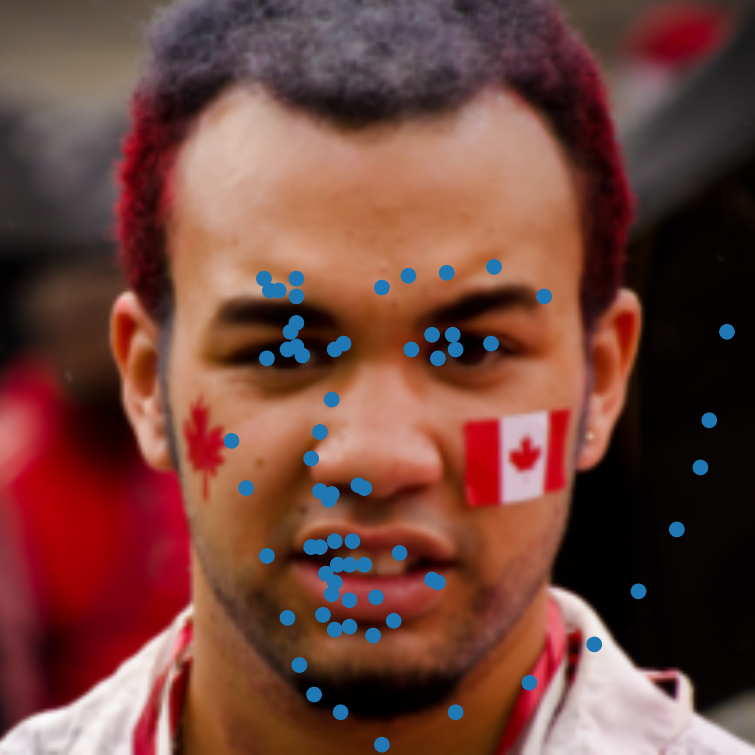}
    \includegraphics[width=0.3\textwidth]{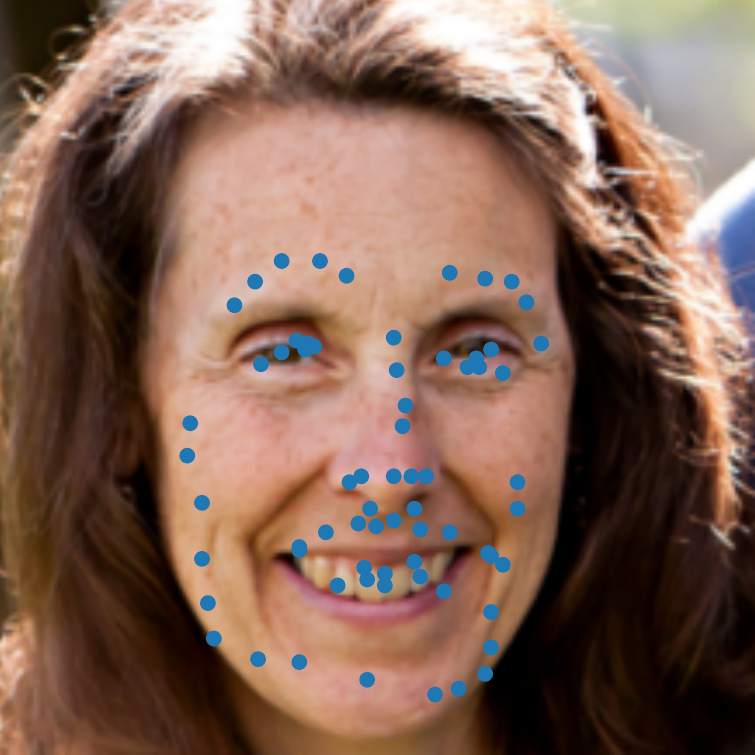}
    \includegraphics[width=0.3\textwidth]{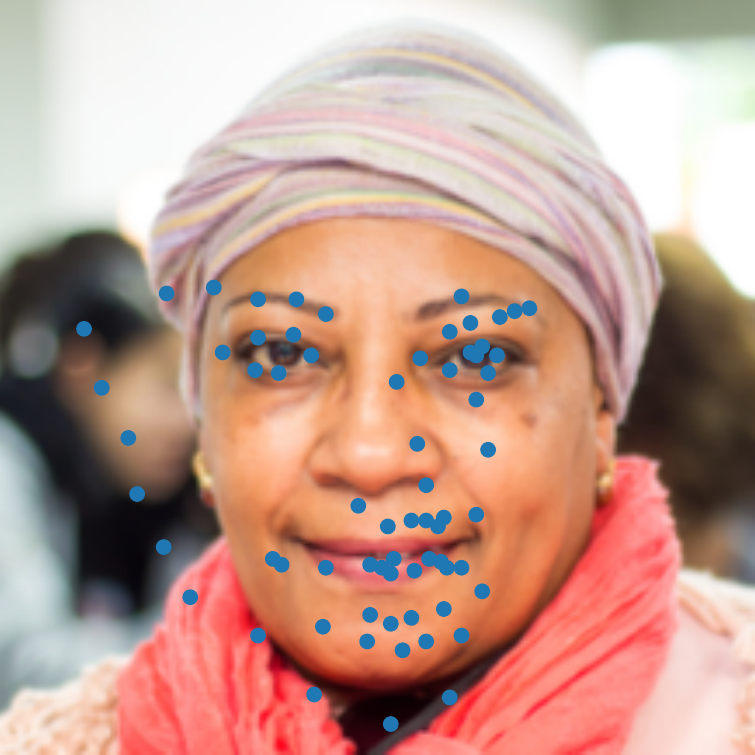}
    \includegraphics[width=0.3\textwidth]{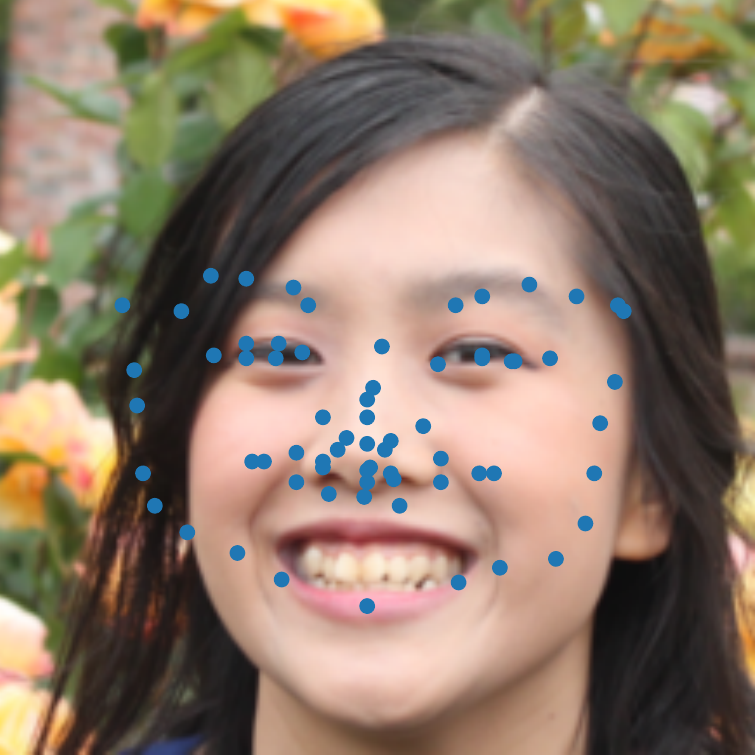}
    \includegraphics[width=0.3\textwidth]{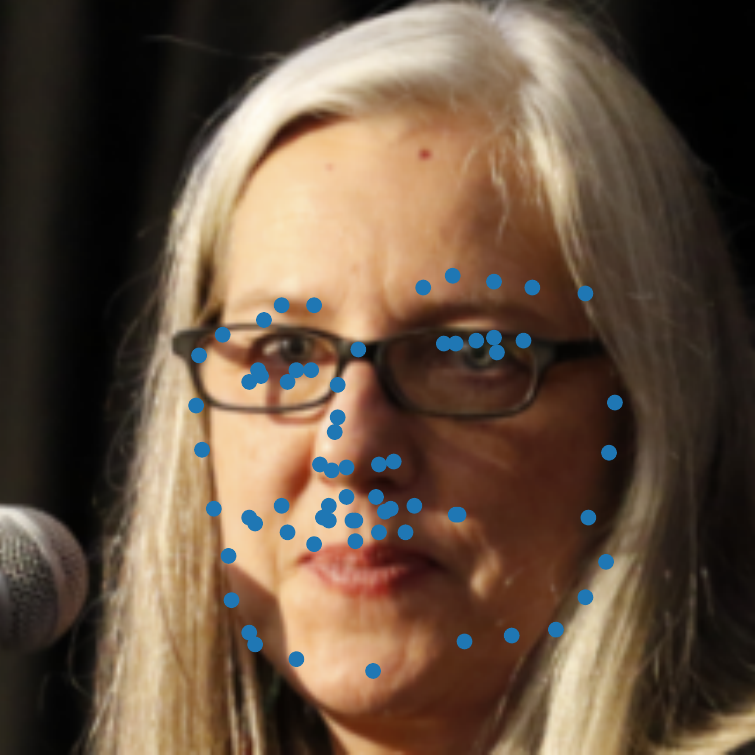}
    \includegraphics[width=0.3\textwidth]{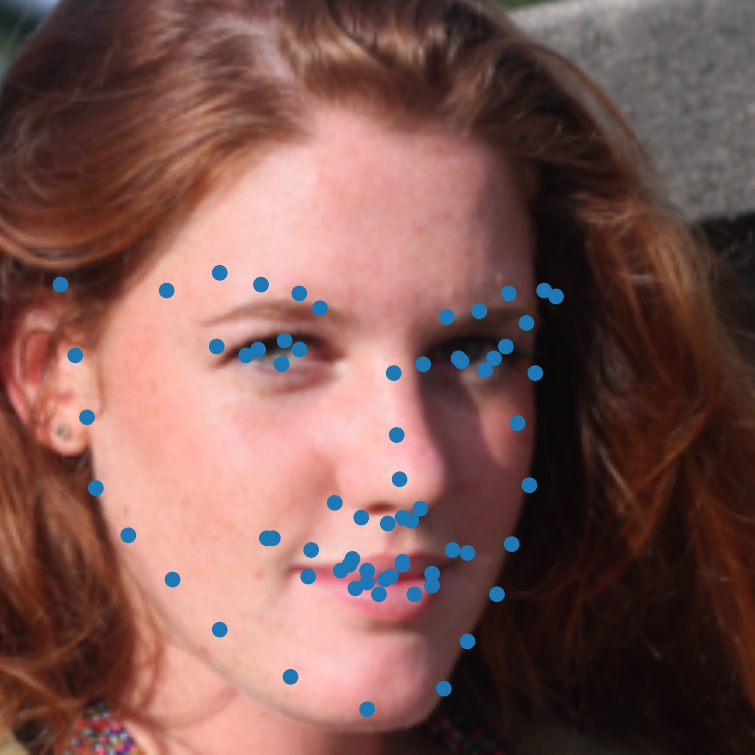}
    \includegraphics[width=0.3\textwidth]{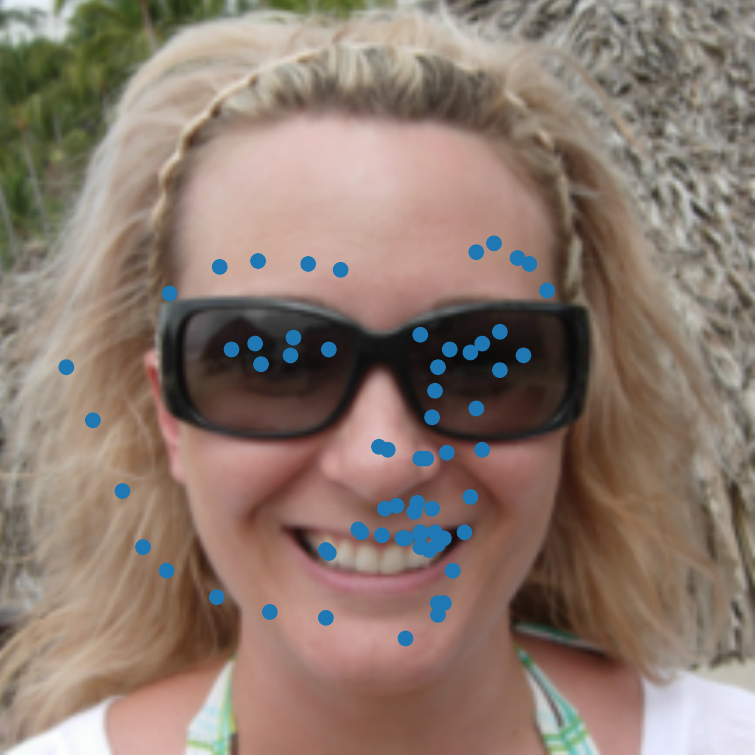}
    \caption{StyleGAN2 generator}
\end{subfigure}
\hfill
\begin{subfigure}[b]{0.49\linewidth}
    \centering
    \includegraphics[width=0.3\textwidth]{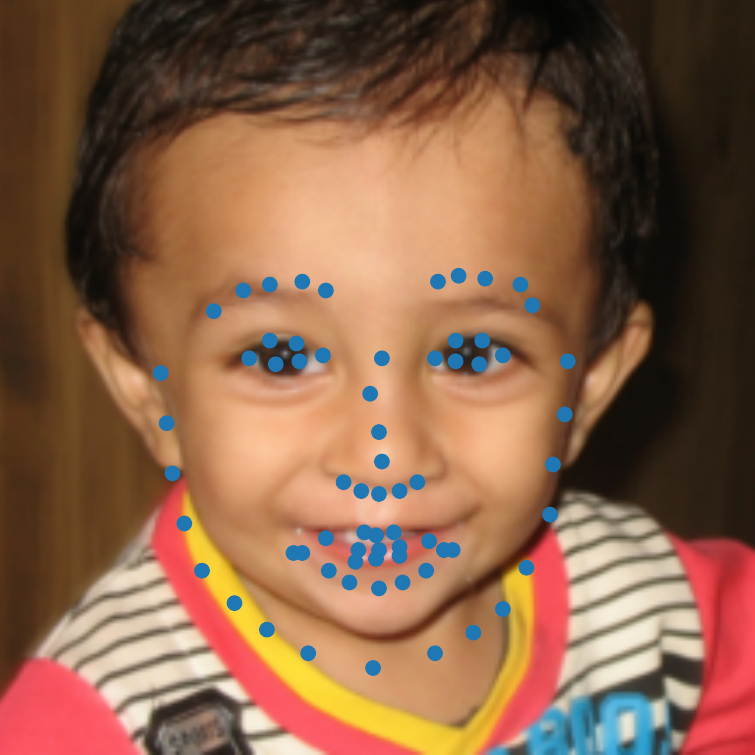}
    \includegraphics[width=0.3\textwidth]{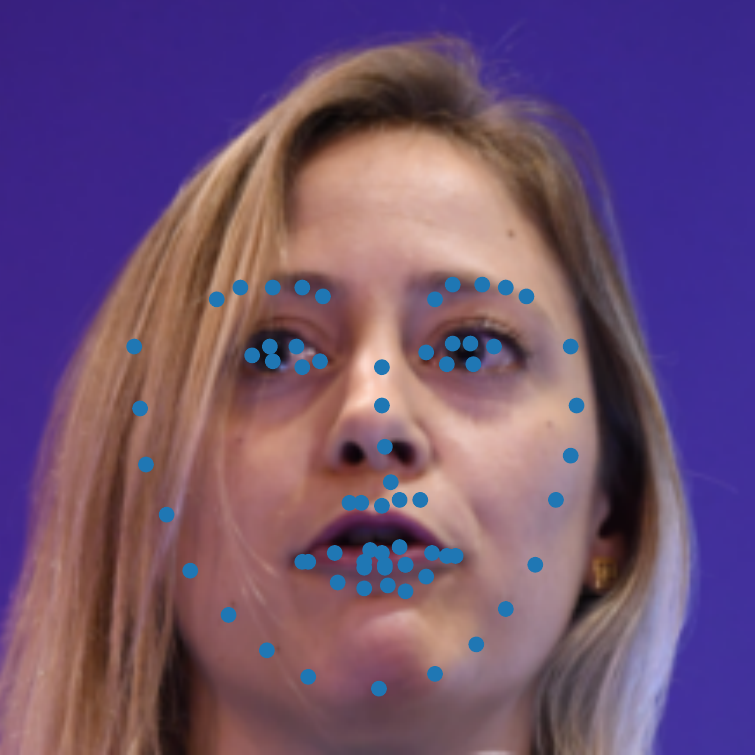}
    \includegraphics[width=0.3\textwidth]{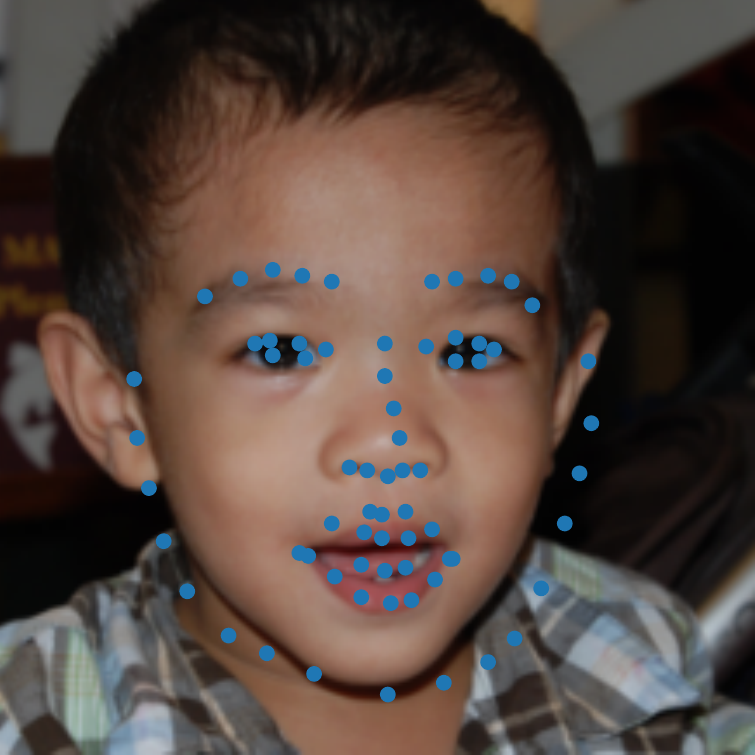}
    \includegraphics[width=0.3\textwidth]{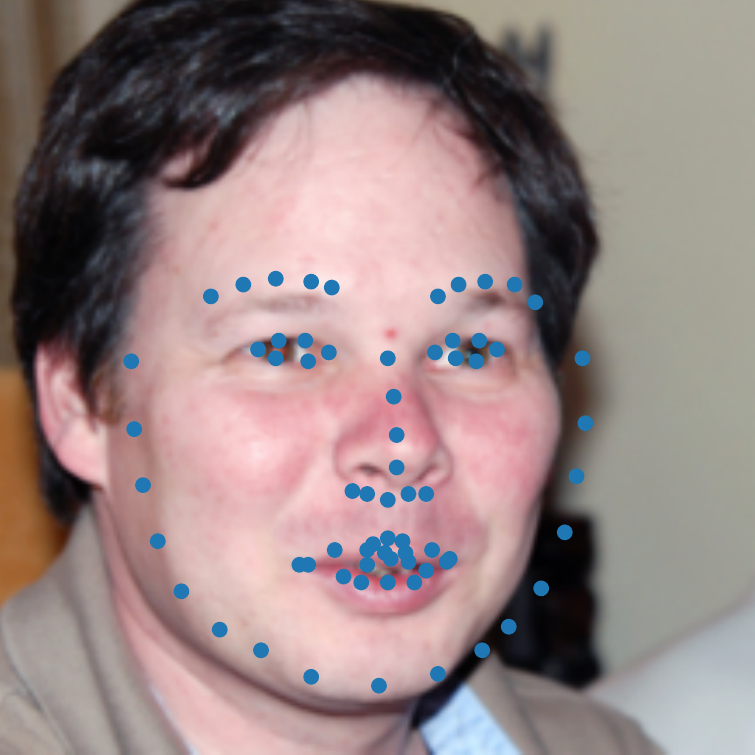}
    \includegraphics[width=0.3\textwidth]{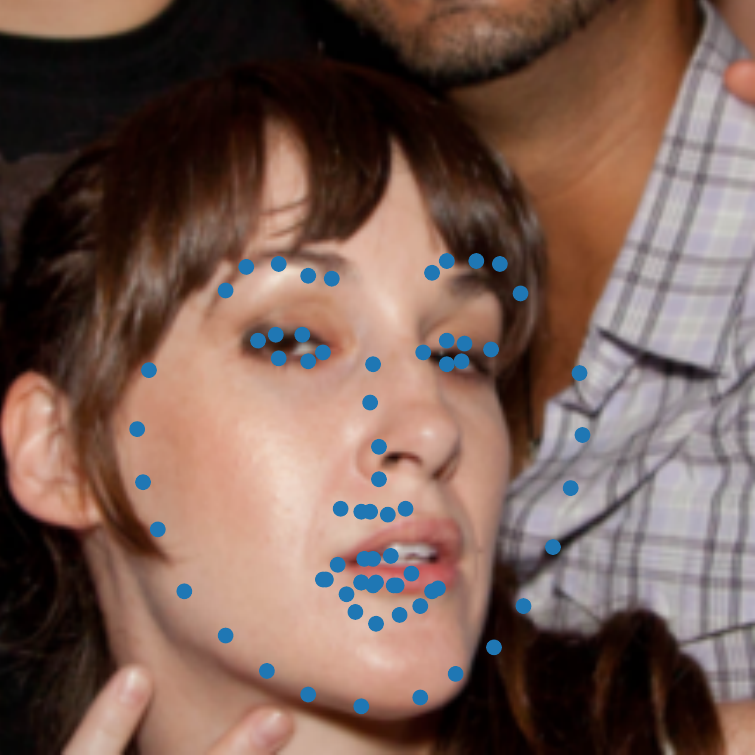}
    \includegraphics[width=0.3\textwidth]{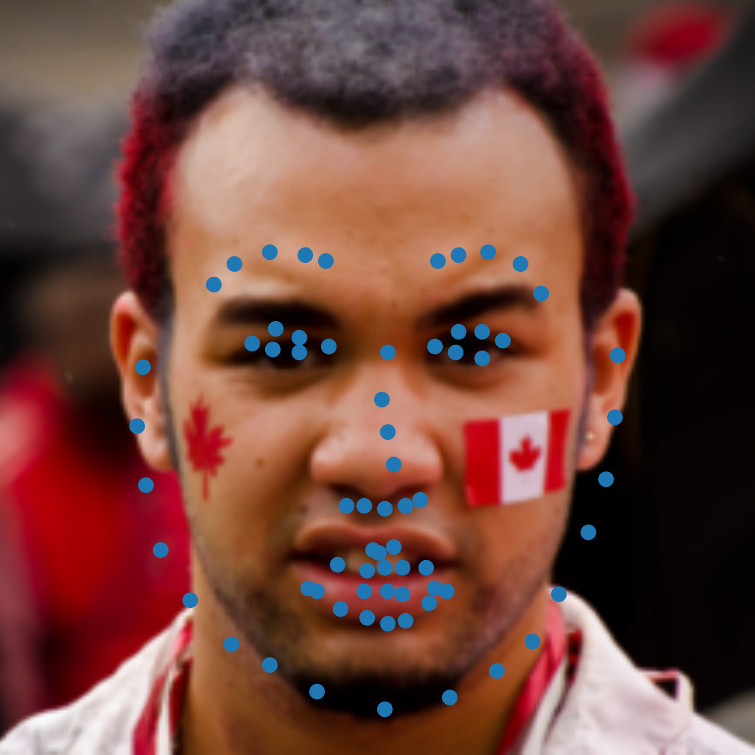}
    \includegraphics[width=0.3\textwidth]{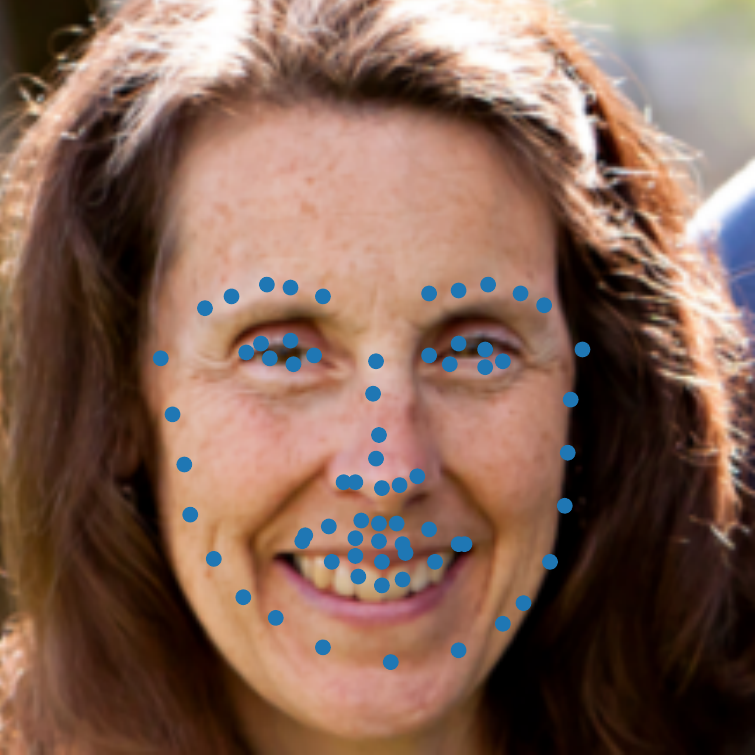}
    \includegraphics[width=0.3\textwidth]{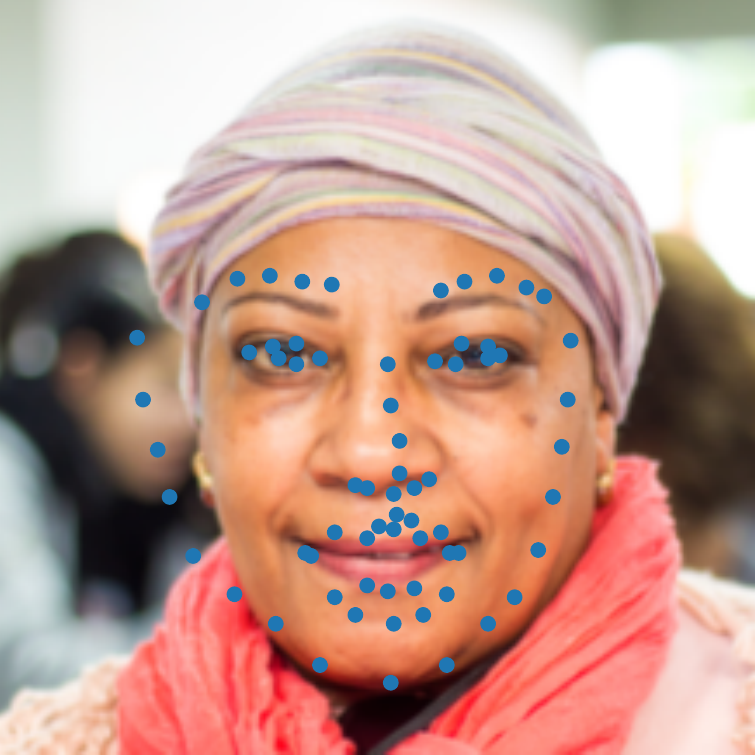}
    \includegraphics[width=0.3\textwidth]{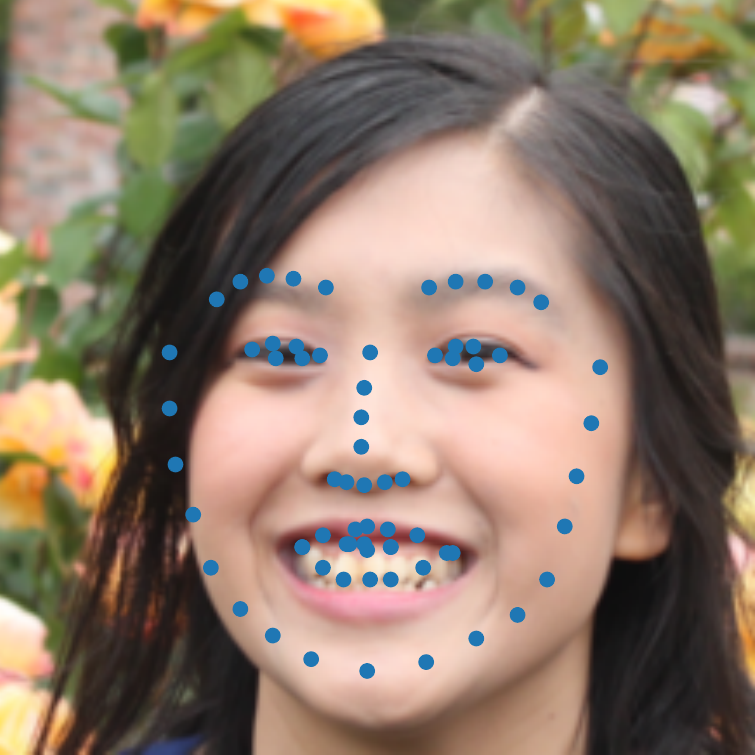}
    \includegraphics[width=0.3\textwidth]{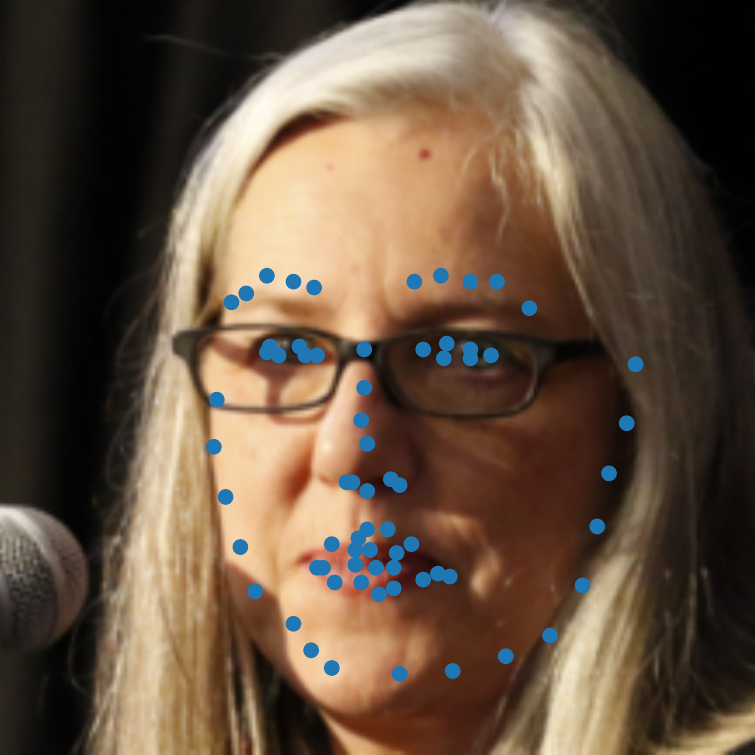}
    \includegraphics[width=0.3\textwidth]{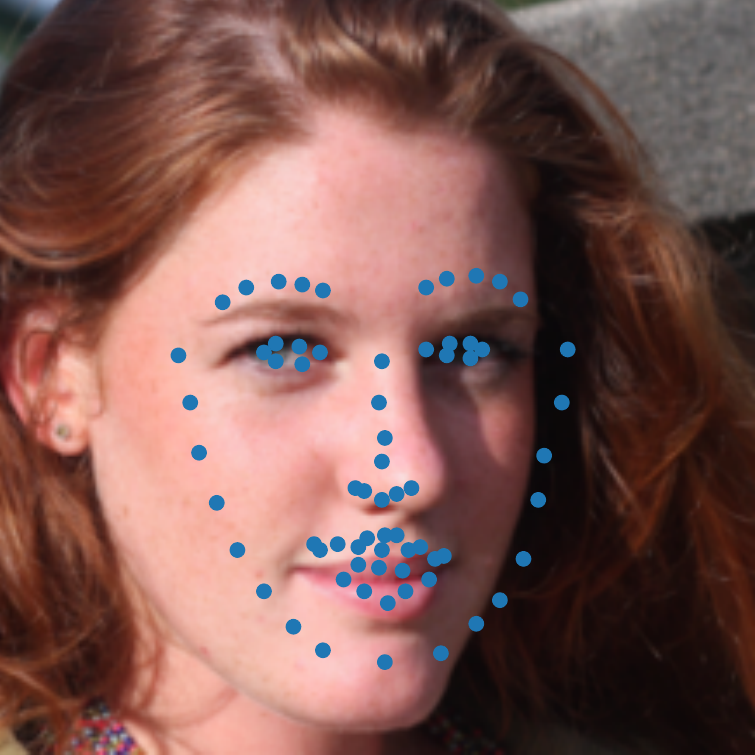}
    \includegraphics[width=0.3\textwidth]{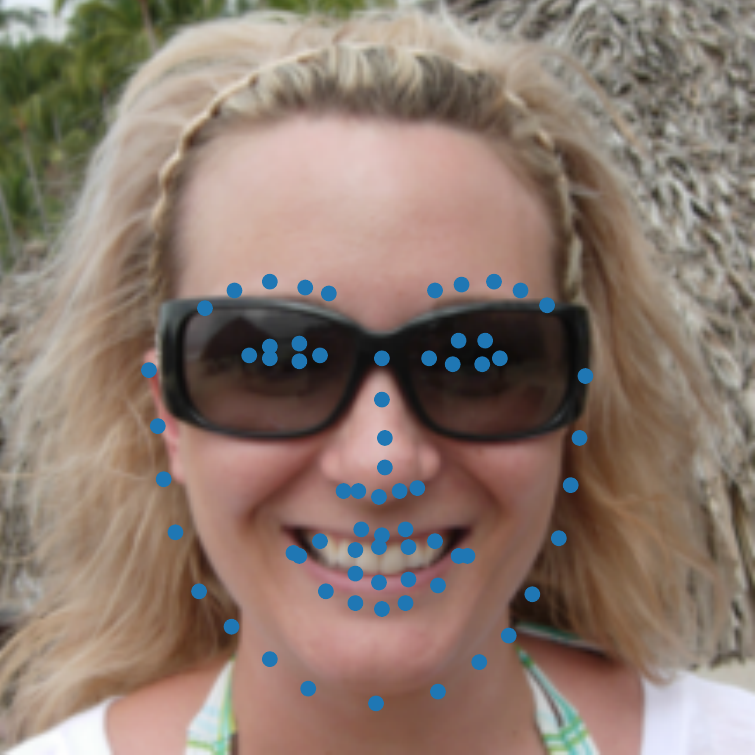}
    \caption{Our INR-based generator}
\end{subfigure}
    \caption{Predicting keypoints from latent codes for random FFHQ images. The corresponding scores are presented in Table~\ref{table:keypoints-prediction}.}
    \label{fig:keypoints-extra}
\end{figure*}

We project FFHQ images into a latent space using the latent space projection procedure from the official repo\footnote{https://github.com/NVlabs/stylegan2/blob/master/projector.py}.
We used default hyperparameters except for it, except that we didn't optimize the injected noise since this would take away some of the information that is better to be stored in the latent code.
We depict additional qualitative results for random samples of FFHQ on \figref{fig:keypoints-extra}.

We provide the algorithm on KPL computation in Algorithm~\ref{alg:kpl}.
We use $N_\text{tr} = 10^4$ and $N_\text{ts} = 256$.

\begin{figure*}
    \centering
    \begin{subfigure}[b]{0.45\linewidth}
        \centering
        \includegraphics[width=\linewidth]{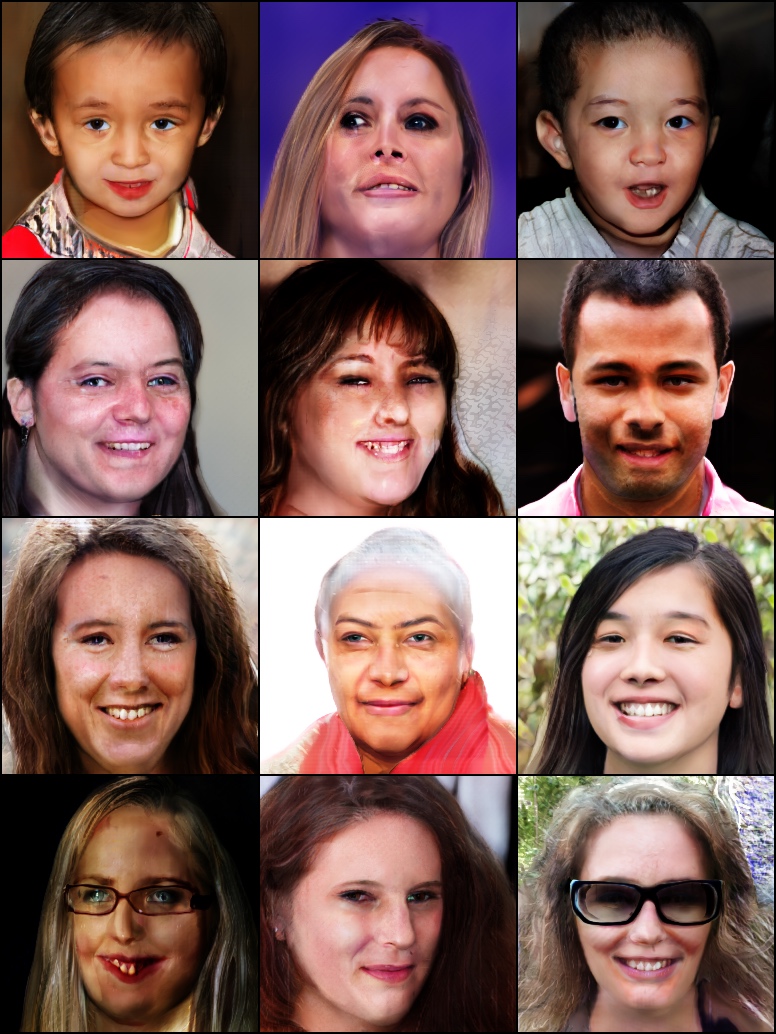}
        \caption{INR-GAN latent space projections.}
    \end{subfigure}
    \hfill
    \begin{subfigure}[b]{0.45\linewidth}
        \centering
        \includegraphics[width=\linewidth]{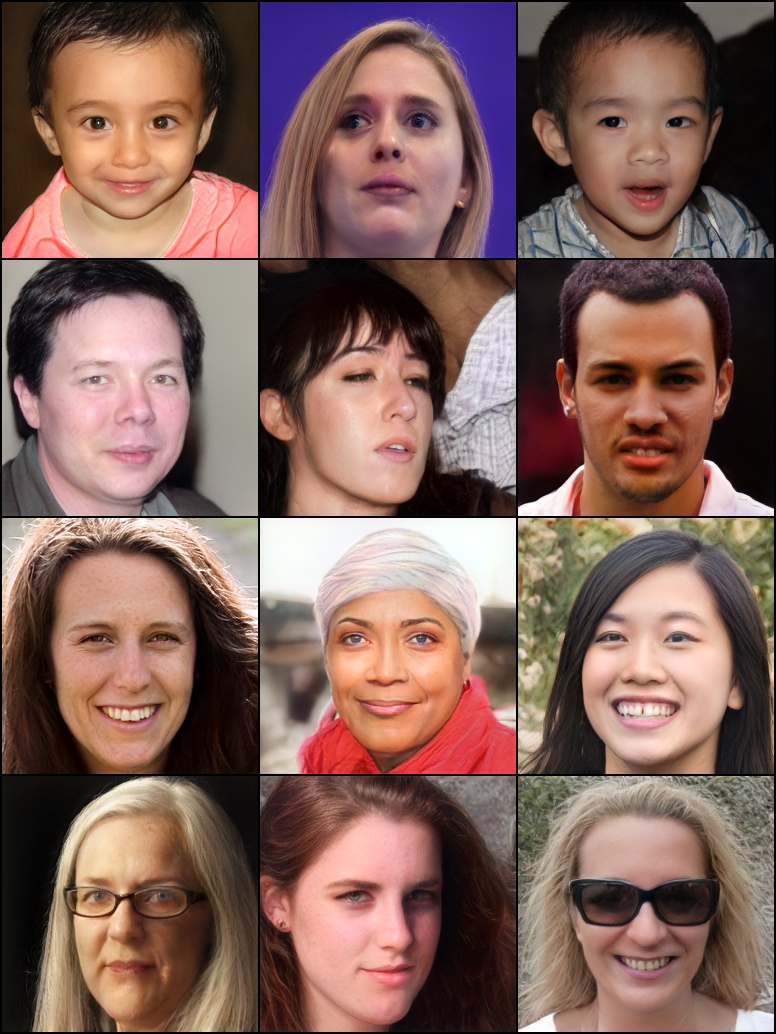}
        \caption{StyleGAN2 latent space projections.}
    \end{subfigure}
    \caption{Projection results by projection images from \ref{fig:keypoints-extra}. We use the original StyleGAN2's projection procedure \cite{StyleGAN2} to project FFHQ dataset images into the latent space of a generator. All low-frequency details, together with the keypoints are reconstructed well. In our case, the reconstruction quality is lower because we do not optimize for spatial noise as StyleGAN2 does because our vanilla INR-GAN architecture does not use spatial noise injection.}
    \label{fig:projection-samples}
\end{figure*}

\begin{algorithm}
\normalsize
\SetAlgoLined
\SetKwInOut{Input}{Input}\SetKwInOut{Output}{Output}
\Input{Keypoints extractor $\mathcal{K}:\bm{x}\mapsto \bm k \in \R^{d_k}$.}
\Input{Generator model $\G:\bm{w} \mapsto \bm{x}$}
\Input{Embedding procedure $\mathcal{E}:\bm{x}\mapsto\bm{w}$.}
\Input{Collection of real face images $X_\text{test} = \{\bm{x}_i\}_{i=1}^n$ of size $N_\text{ts}$.}
\Output{KPL score $s \in [0, +\infty)$.}

Generate $N_\text{tr}$ latent codes $W_\text{train} = \{\bm{w}_1, ..., \bm{w}_{N_\text{tr}}\}$\;
Generate a dataset of synthetic images $X_\text{train} = \{ \G(\bm w_i) | \bm w_i \in W_\text{train} \}$\;
Extract keypoints $K_\text{train} = \{\mathcal{K}(\bm x_i) ~|~ \bm x_i \in X_\text{train}\}$ and $K_\text{test} = \{\mathcal{K}(\bm x_j) ~|~ \bm x_j \in X_\text{test}\}$\;
Embed real images $W_\text{test} = \{\mathcal{E}(\bm x_j) ~|~ \bm x_j \in X_\text{test}\}$\;
Train a linear keypoints estimator $(\bm A^*, \bm b^*) = \arg\min_{\bm A,\bm b} \sum_{i=1}^{N_\text{tr}} \| (\bm{A}\bm{w}_i + \bm{b}) - \bm{k}_i\|_2^2$\;
Evaluate its performance on the test set: $s = \sum_{i=1}^{N_\text{ts}} \| (\bm{A}\bm{w}_j + \bm{b}) - \bm{k}_j\|_2^2$\;
Return $s$\;
\caption{Compute Keypoints Prediction Loss (KPL).}
\label{alg:kpl}
\end{algorithm}

\subsection{Additional samples}
\begin{figure*}
    \centering
    \begin{subfigure}[b]{0.48\linewidth}
        \centering
        \includegraphics[width=\textwidth]{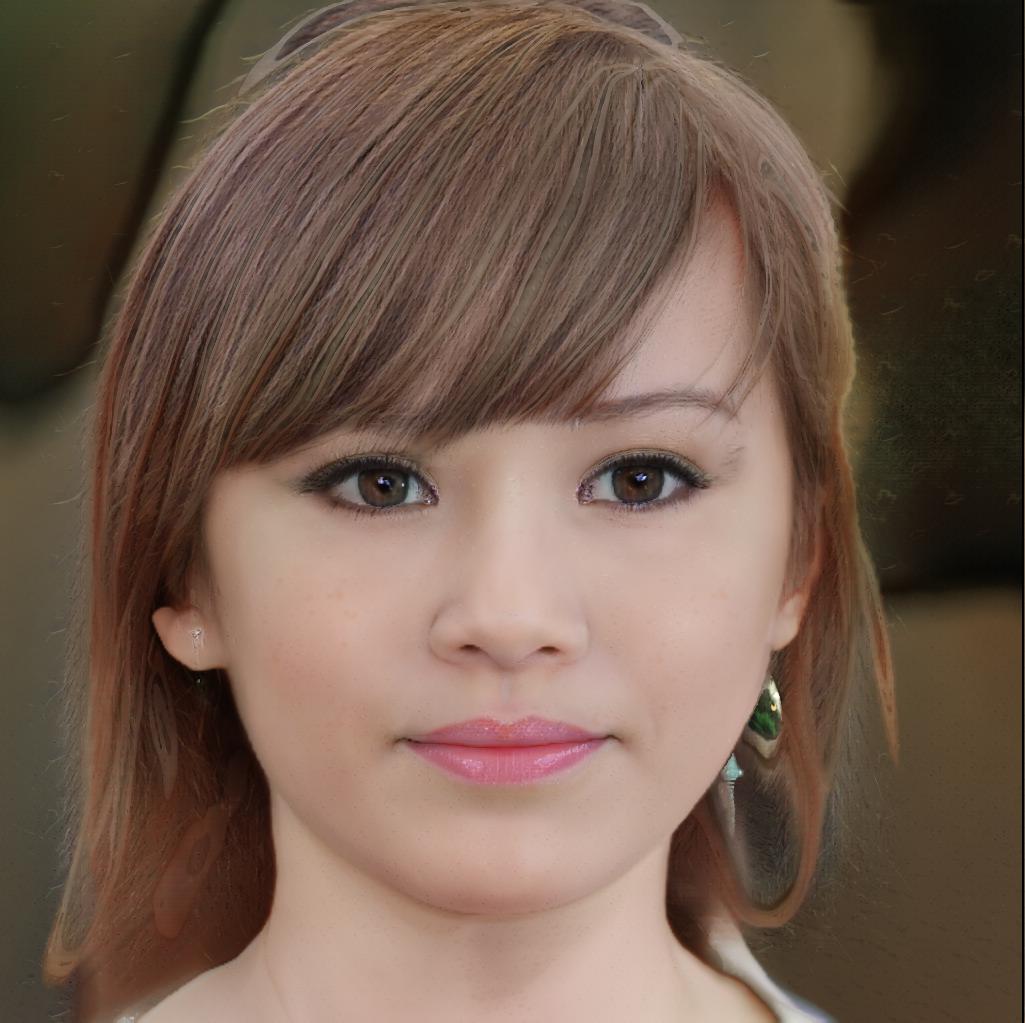}
    \end{subfigure}
    \hfill
    \begin{subfigure}[b]{0.49\linewidth}
        \centering
        \begin{subfigure}[b]{0.49\linewidth}
        \centering
        \includegraphics[width=\textwidth]{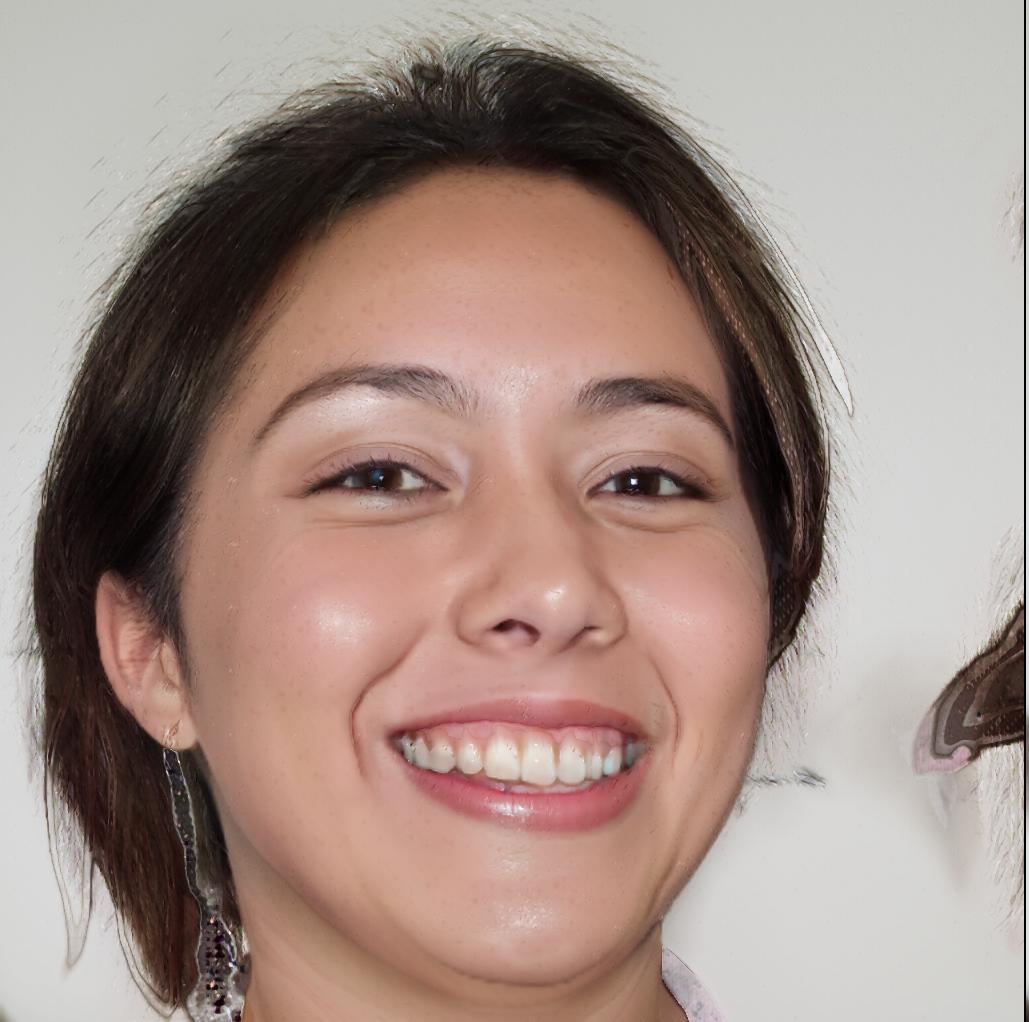}
        \end{subfigure}
        \hfill
        \begin{subfigure}[b]{0.49\linewidth}
        \centering
        \includegraphics[width=\textwidth]{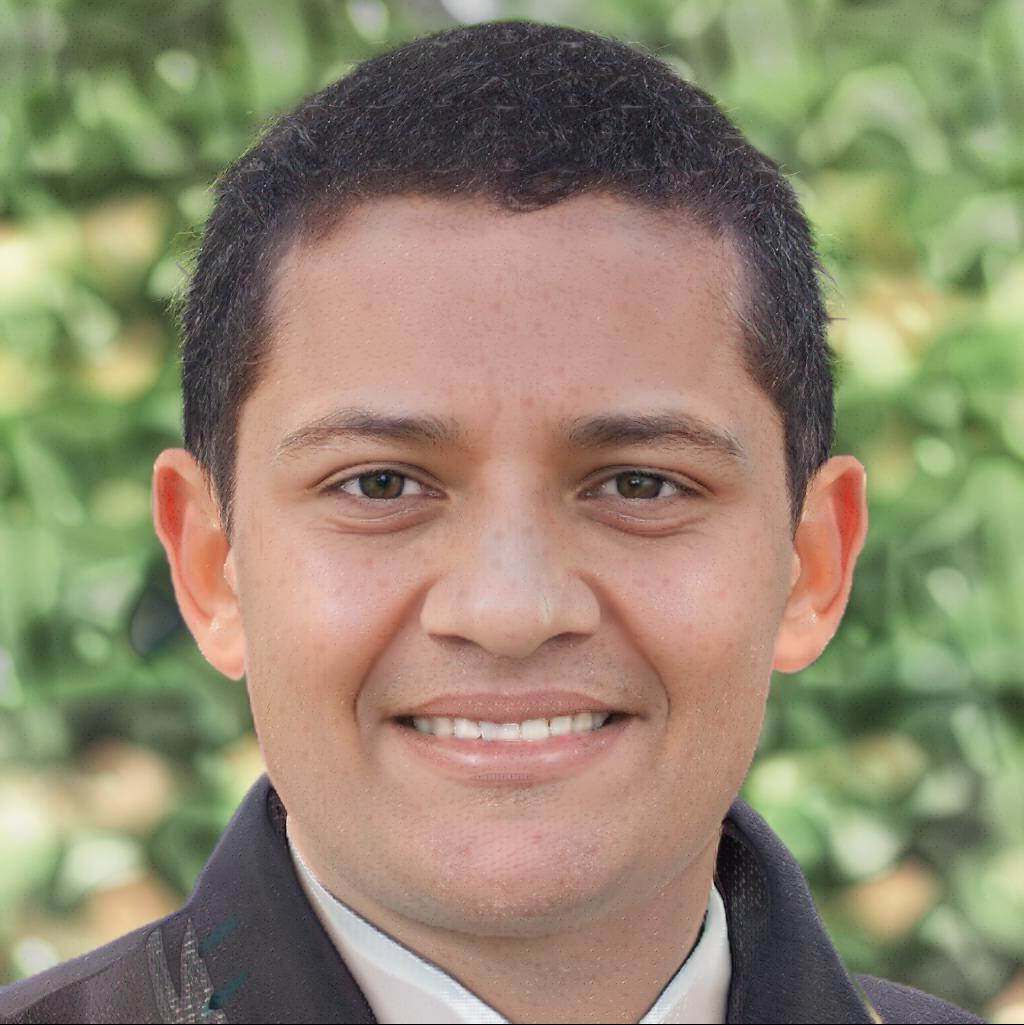}
        \end{subfigure}
        \vfill
        \begin{subfigure}[b]{0.49\linewidth}
        \centering
        \includegraphics[width=\textwidth]{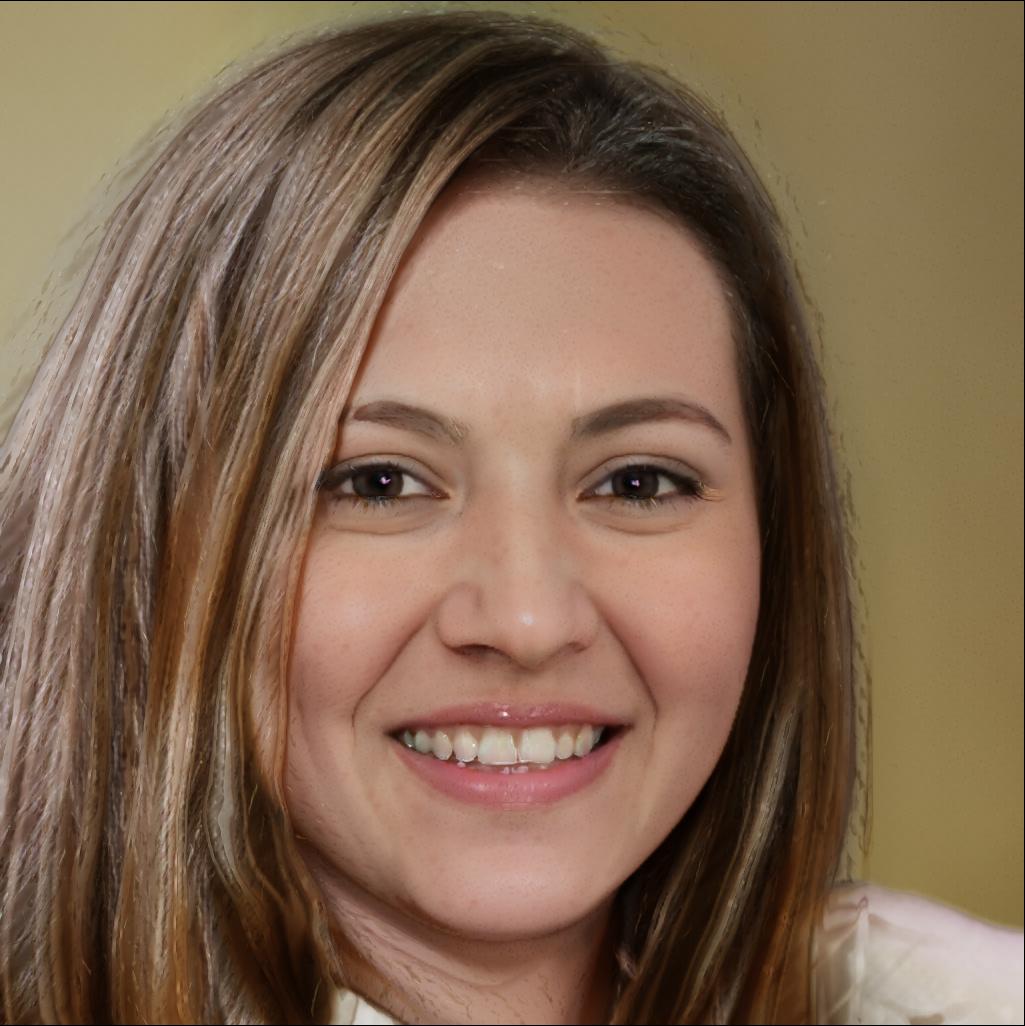}
        \end{subfigure}
        \hfill
        \begin{subfigure}[b]{0.49\linewidth}
        \centering
        \includegraphics[width=\textwidth]{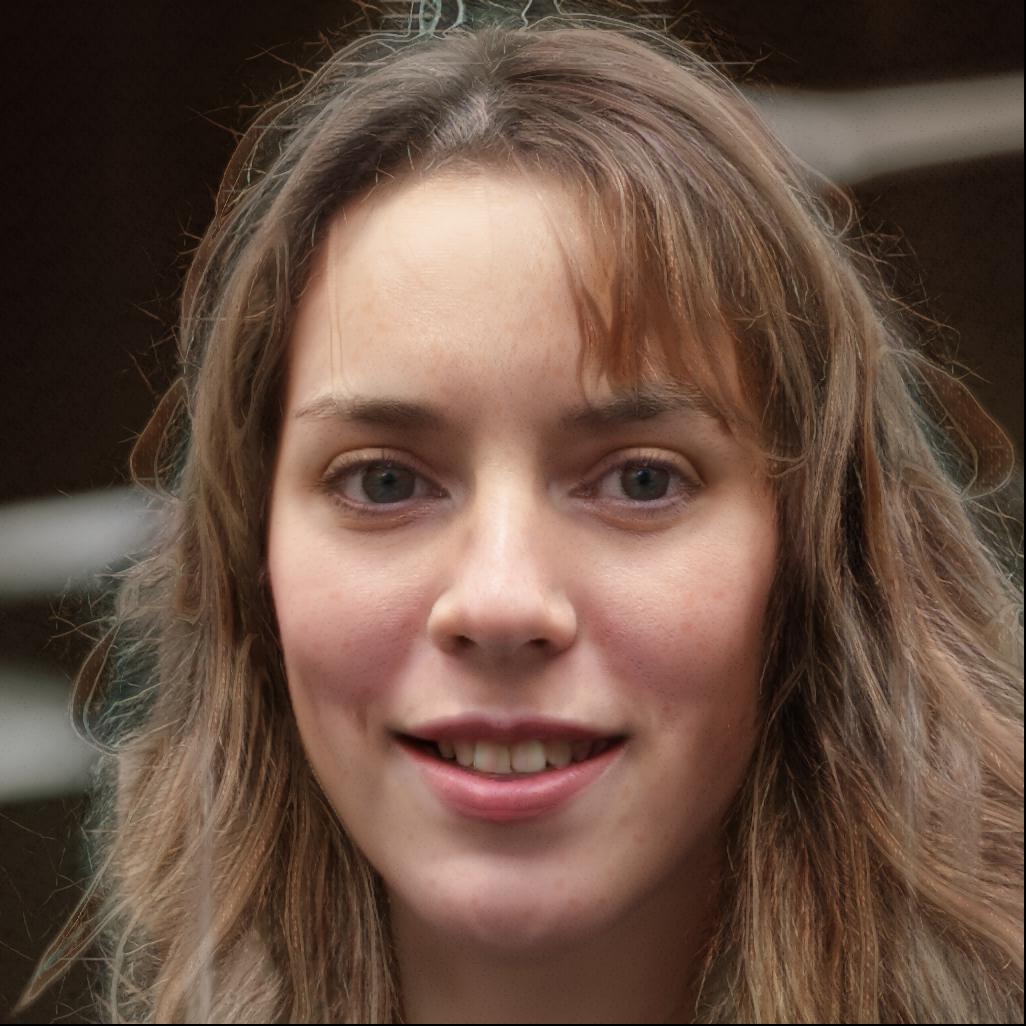}
        \end{subfigure}
    \end{subfigure}
    \caption{Random (uncurated) samples from our model trained on FFHQ $1024 \times 1024$ dataset with the truncation factor of 0.9. FID: 16.32}
    \label{fig:ffhq1024-samples}
\end{figure*}

\begin{figure*}
    \centering
    \includegraphics[width=0.24\textwidth]{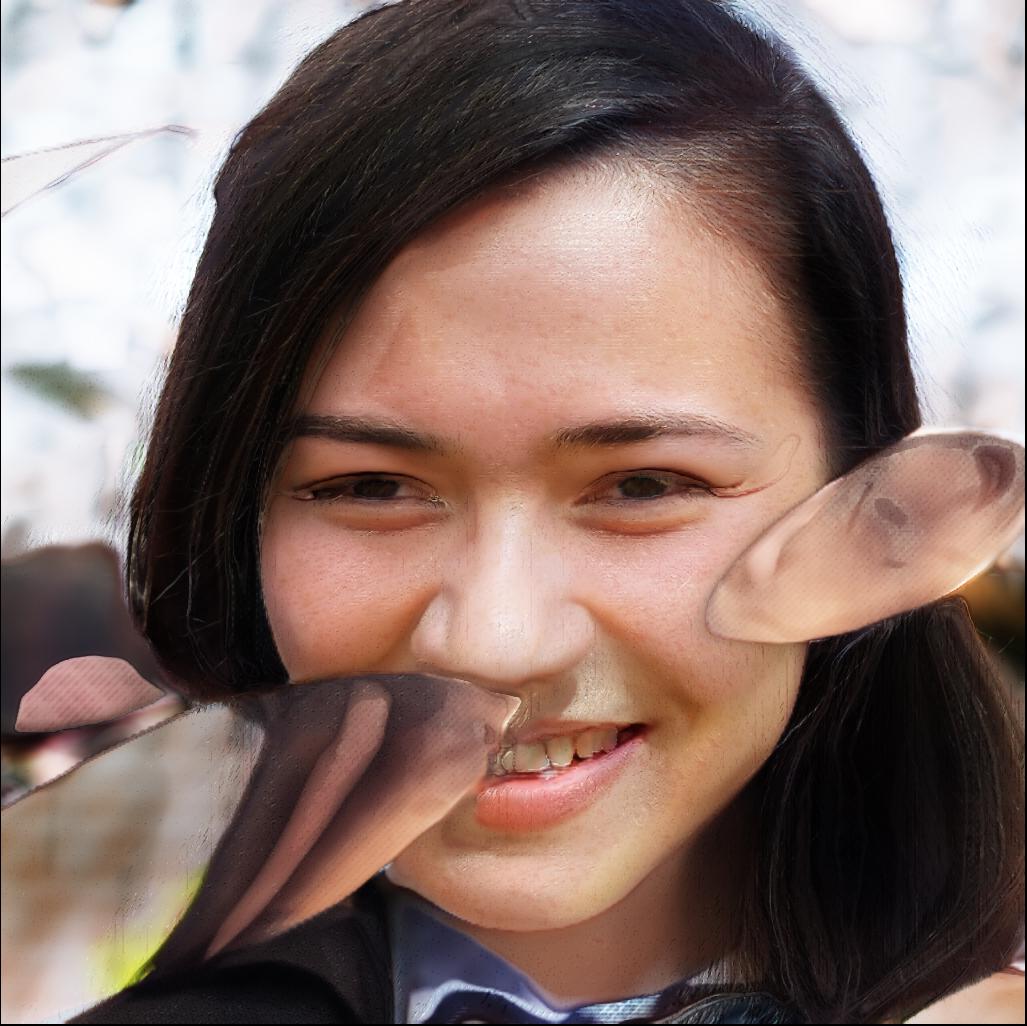}
    \includegraphics[width=0.24\textwidth]{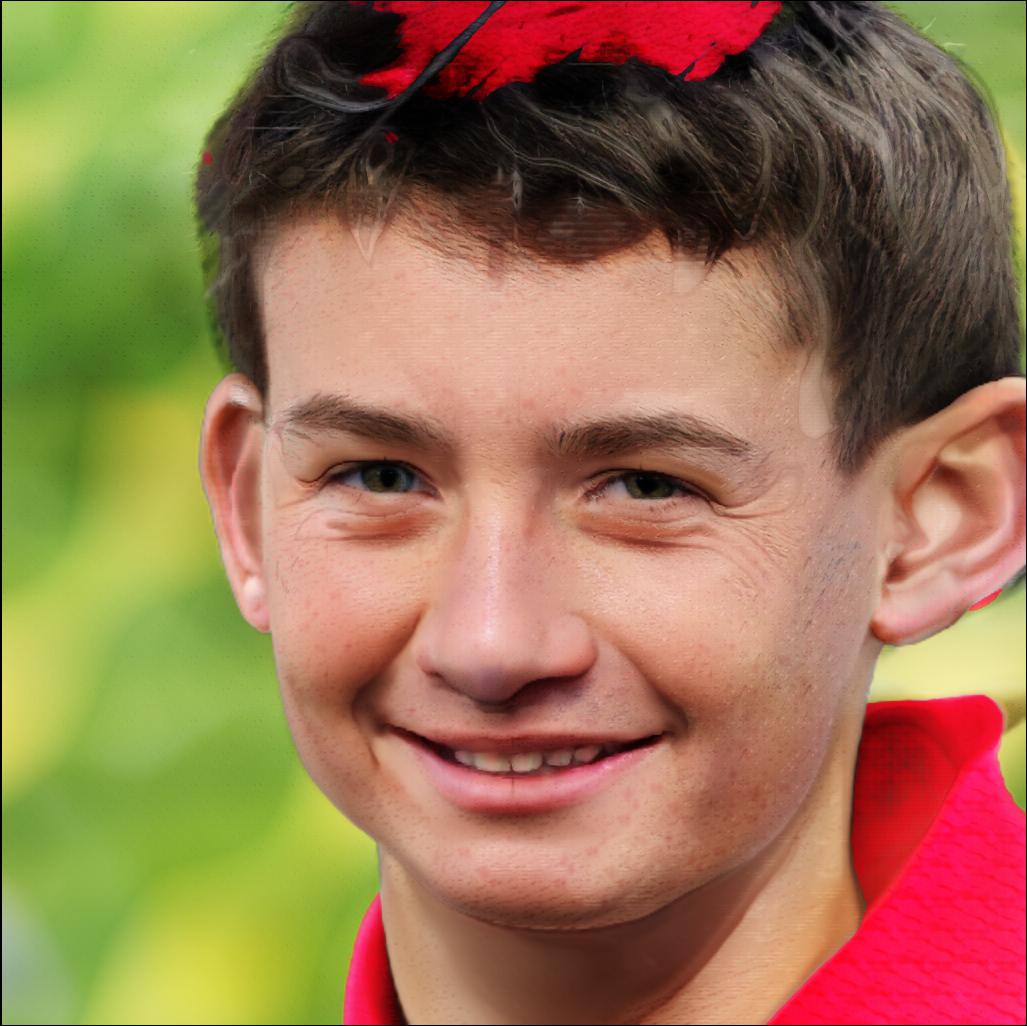}
    \includegraphics[width=0.24\textwidth]{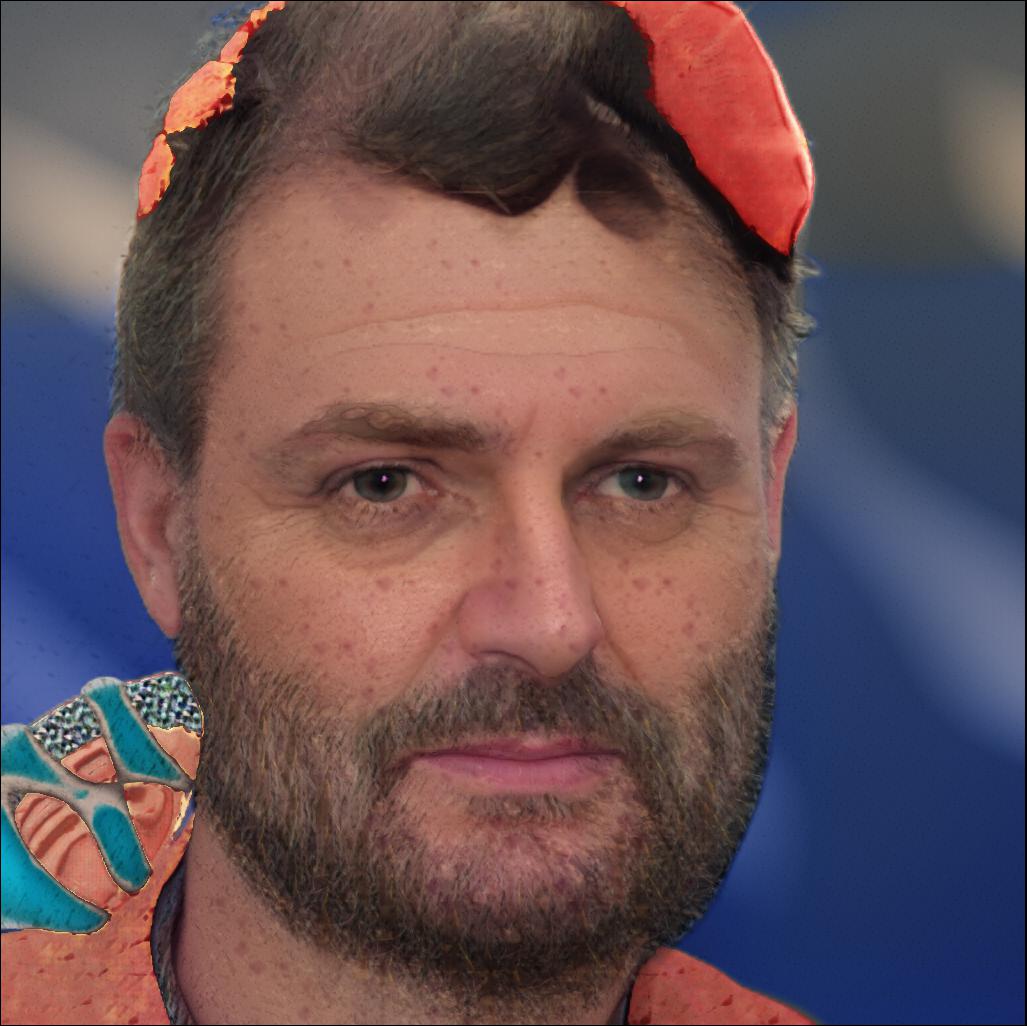}
    \includegraphics[width=0.24\textwidth]{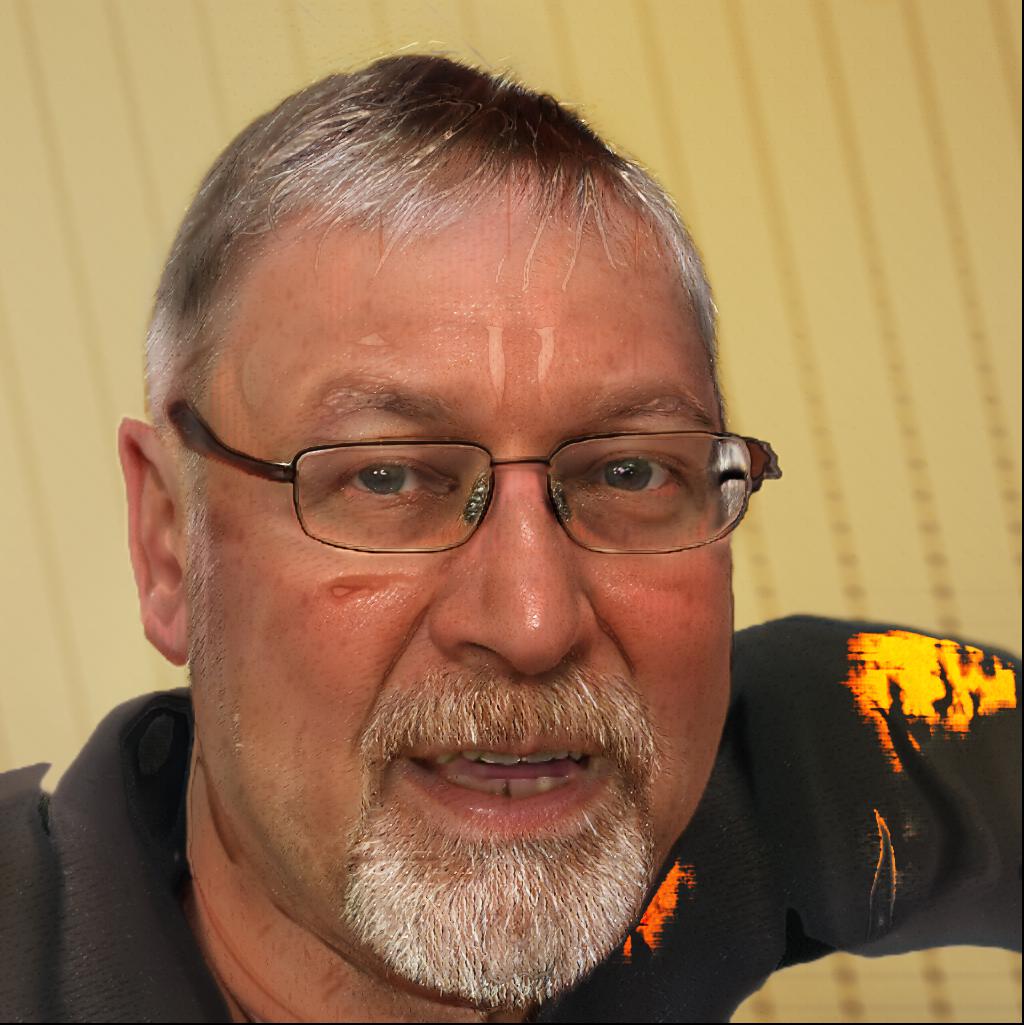}
    \caption{Common artifacts found in our model's samples when sampling without truncation. As one can see, the most severe ones are ``stains'' and patterned texture.}
    \label{fig:ffhq1024-artifacts}
\end{figure*}

\begin{figure*}
    \captionsetup[subfigure]{labelformat=empty}
    \centering

\begin{subfigure}[b]{0.8\textwidth}
    \begin{subfigure}[b]{0.1\linewidth}
        \centering
        \includegraphics[width=\textwidth]{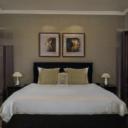}
    \end{subfigure}
    \begin{subfigure}[b]{0.22\linewidth}
        \centering
        \includegraphics[width=\textwidth]{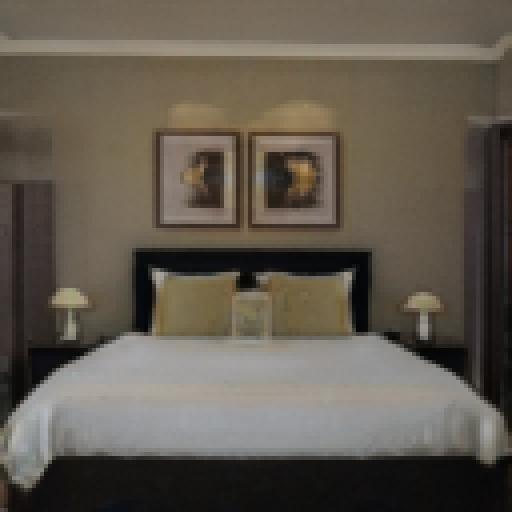}
    \end{subfigure}
    \begin{subfigure}[b]{0.22\linewidth}
        \centering
        \includegraphics[width=\textwidth]{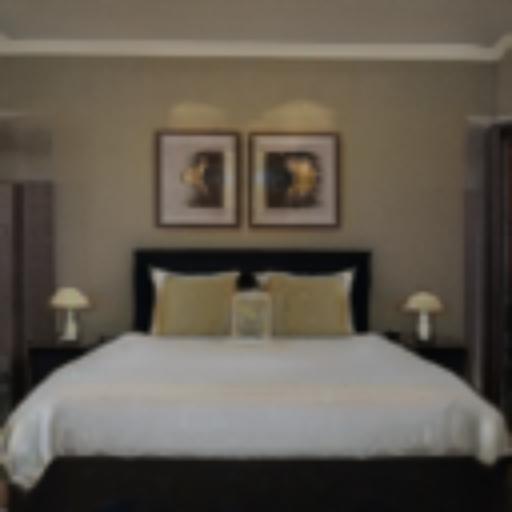}
    \end{subfigure}
    \begin{subfigure}[b]{0.22\linewidth}
        \centering
        \includegraphics[width=\textwidth]{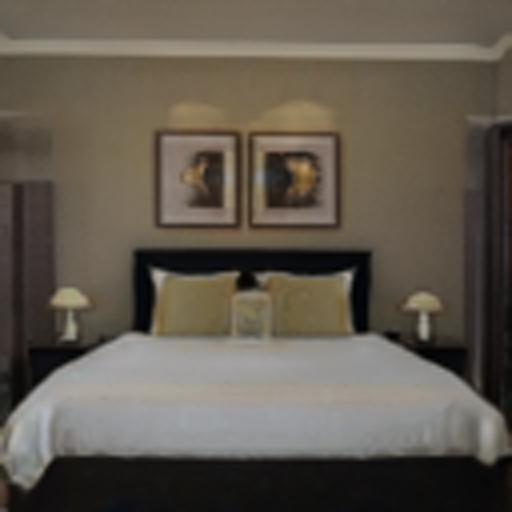}
    \end{subfigure}
    \begin{subfigure}[b]{0.22\linewidth}
        \centering
        \includegraphics[width=\textwidth]{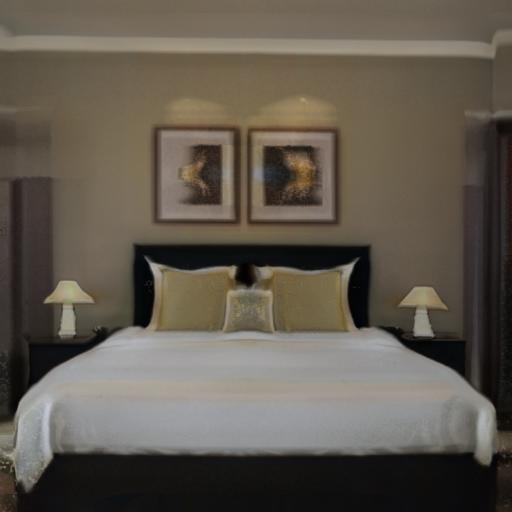}
    \end{subfigure}
\end{subfigure}
\begin{subfigure}[b]{0.8\textwidth}
    \begin{subfigure}[b]{0.1\linewidth}
        \centering
        \includegraphics[width=\textwidth]{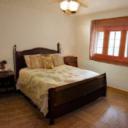}
    \end{subfigure}
    \begin{subfigure}[b]{0.22\linewidth}
        \centering
        \includegraphics[width=\textwidth]{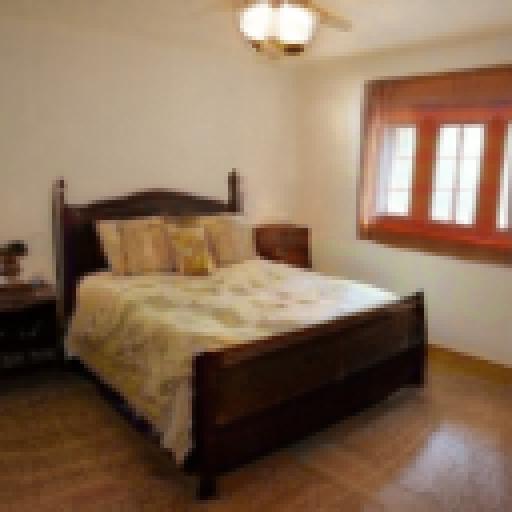}
    \end{subfigure}
    \begin{subfigure}[b]{0.22\linewidth}
        \centering
        \includegraphics[width=\textwidth]{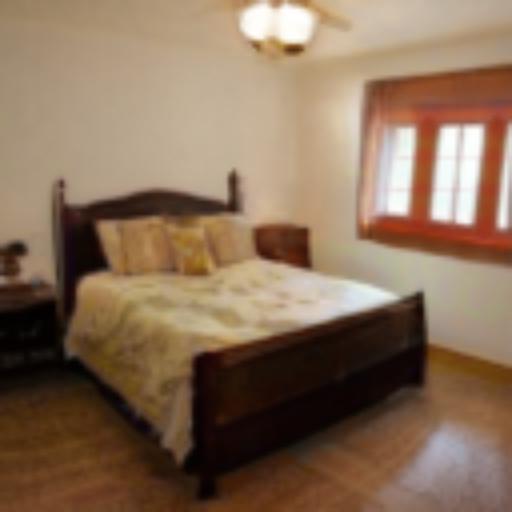}
    \end{subfigure}
    \begin{subfigure}[b]{0.22\linewidth}
        \centering
        \includegraphics[width=\textwidth]{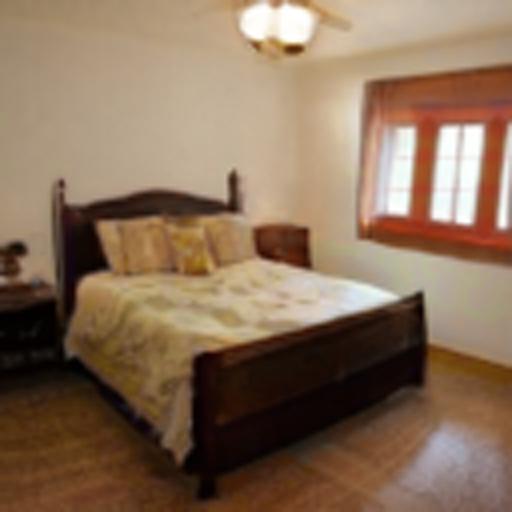}
    \end{subfigure}
    \begin{subfigure}[b]{0.22\linewidth}
        \centering
        \includegraphics[width=\textwidth]{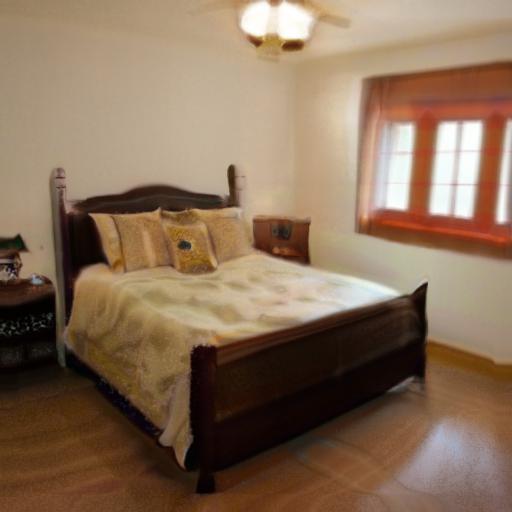}
    \end{subfigure}
\end{subfigure}
\begin{subfigure}[b]{0.8\textwidth}
    \begin{subfigure}[b]{0.1\linewidth}
        \centering
        \includegraphics[width=\textwidth]{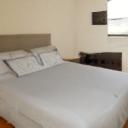}
    \end{subfigure}
    \begin{subfigure}[b]{0.22\linewidth}
        \centering
        \includegraphics[width=\textwidth]{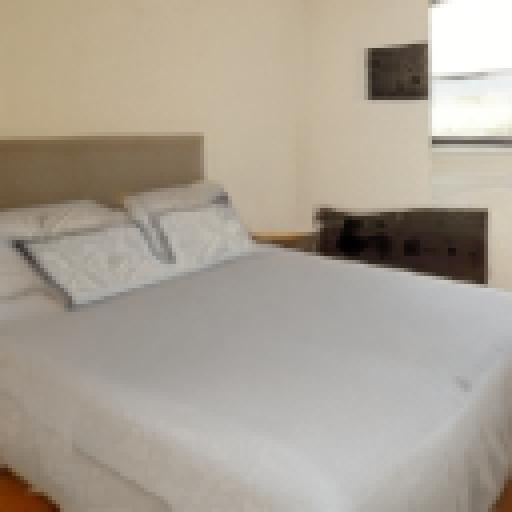}
    \end{subfigure}
    \begin{subfigure}[b]{0.22\linewidth}
        \centering
        \includegraphics[width=\textwidth]{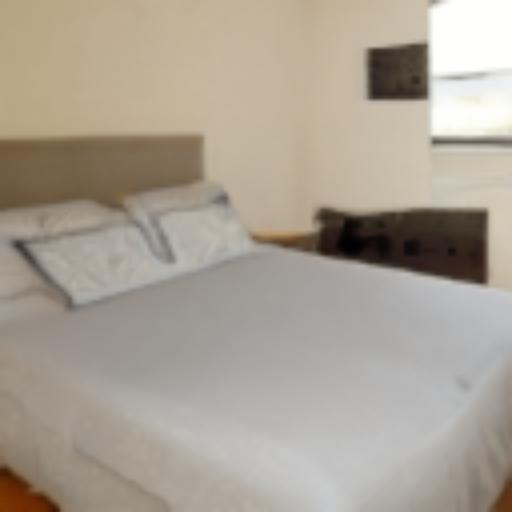}
    \end{subfigure}
    \begin{subfigure}[b]{0.22\linewidth}
        \centering
        \includegraphics[width=\textwidth]{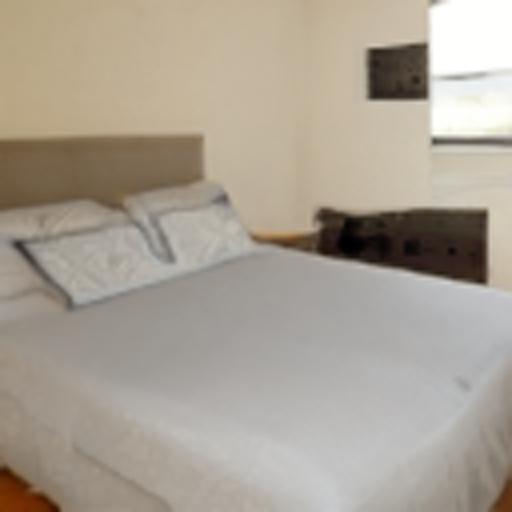}
    \end{subfigure}
    \begin{subfigure}[b]{0.22\linewidth}
        \centering
        \includegraphics[width=\textwidth]{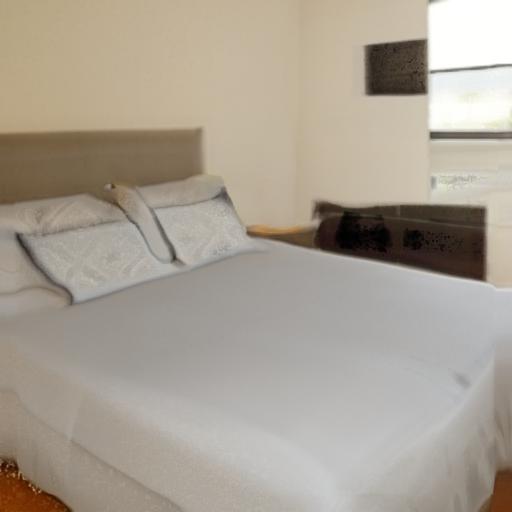}
    \end{subfigure}
\end{subfigure}
\begin{subfigure}[b]{0.8\textwidth}
    \begin{subfigure}[b]{0.1\linewidth}
        \centering
        \includegraphics[width=\textwidth]{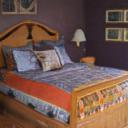}
    \end{subfigure}
    \begin{subfigure}[b]{0.22\linewidth}
        \centering
        \includegraphics[width=\textwidth]{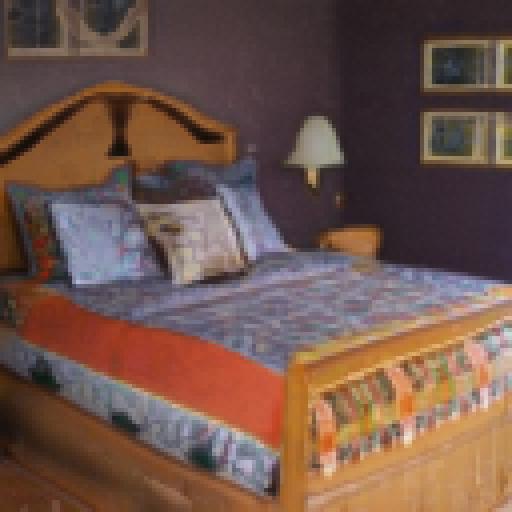}
    \end{subfigure}
    \begin{subfigure}[b]{0.22\linewidth}
        \centering
        \includegraphics[width=\textwidth]{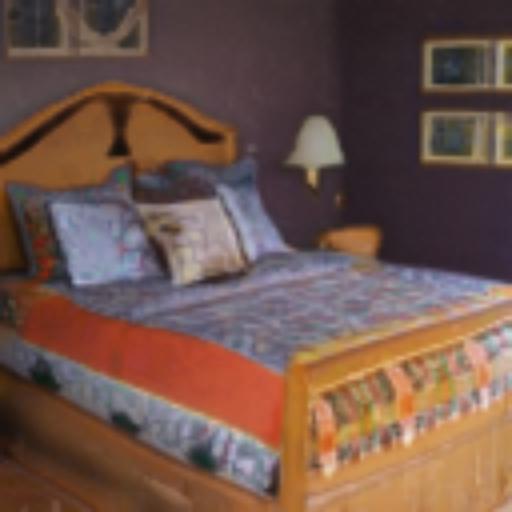}
    \end{subfigure}
    \begin{subfigure}[b]{0.22\linewidth}
        \centering
        \includegraphics[width=\textwidth]{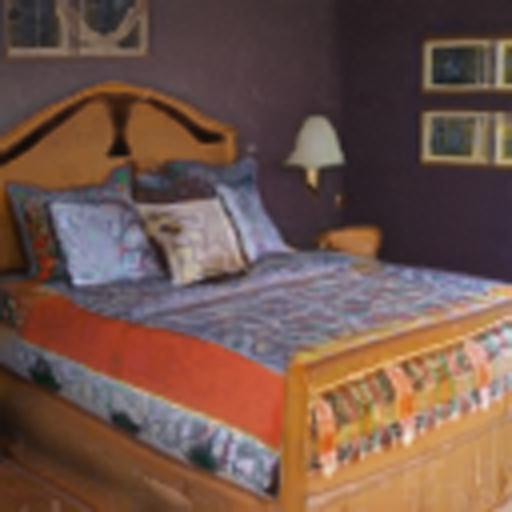}
    \end{subfigure}
    \begin{subfigure}[b]{0.22\linewidth}
        \centering
        \includegraphics[width=\textwidth]{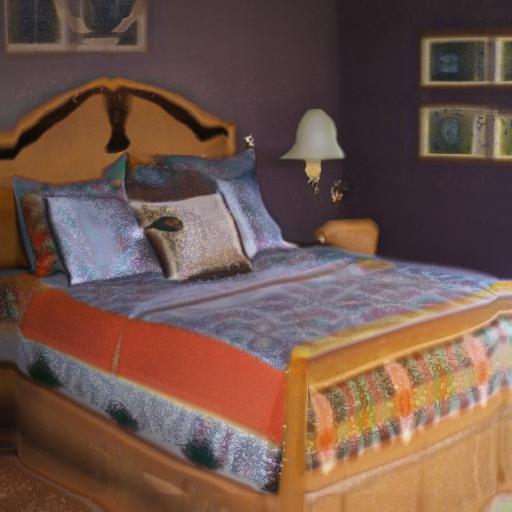}
    \end{subfigure}
\end{subfigure}
\begin{subfigure}[b]{0.8\textwidth}
    \begin{subfigure}[b]{0.1\linewidth}
        \centering
        \includegraphics[width=\textwidth]{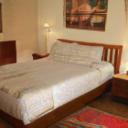}
    \end{subfigure}
    \begin{subfigure}[b]{0.22\linewidth}
        \centering
        \includegraphics[width=\textwidth]{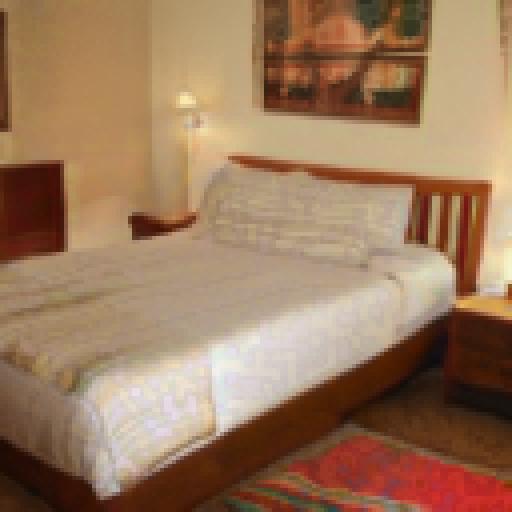}
    \end{subfigure}
    \begin{subfigure}[b]{0.22\linewidth}
        \centering
        \includegraphics[width=\textwidth]{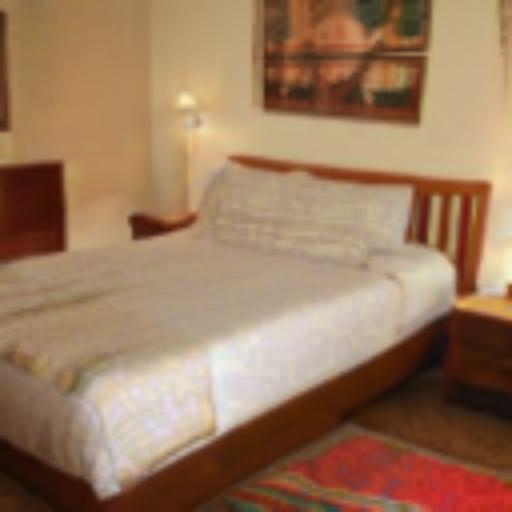}
    \end{subfigure}
    \begin{subfigure}[b]{0.22\linewidth}
        \centering
        \includegraphics[width=\textwidth]{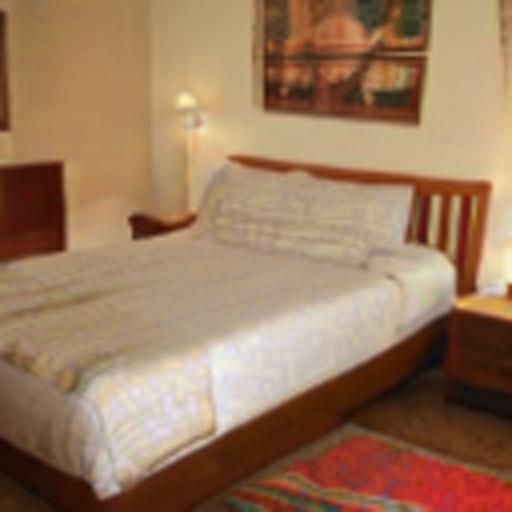}
    \end{subfigure}
    \begin{subfigure}[b]{0.22\linewidth}
        \centering
        \includegraphics[width=\textwidth]{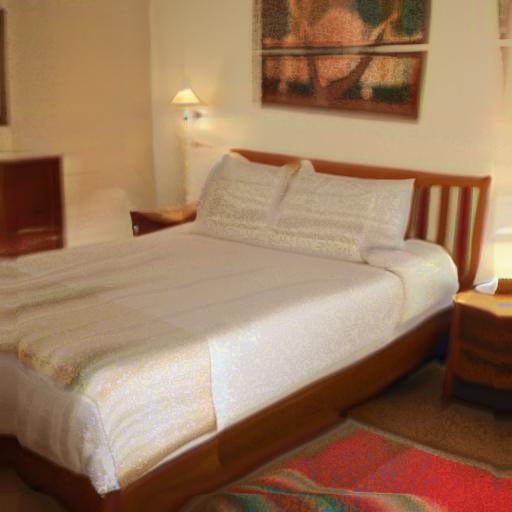}
    \end{subfigure}
\end{subfigure}

\begin{subfigure}[b]{0.8\textwidth}
    \begin{subfigure}[b]{0.1\linewidth}
        \centering
        \includegraphics[width=\textwidth]{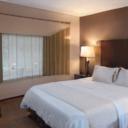}
        \caption{Original}
    \end{subfigure}
    \begin{subfigure}[b]{0.22\linewidth}
        \centering
        \includegraphics[width=\textwidth]{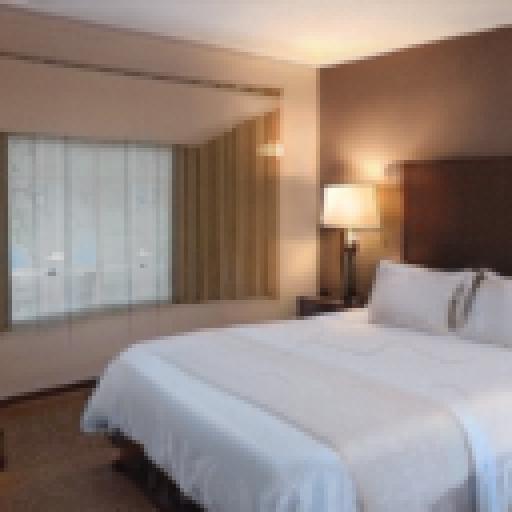}
        \caption{Nearest neighbour}
    \end{subfigure}
    \begin{subfigure}[b]{0.22\linewidth}
        \centering
        \includegraphics[width=\textwidth]{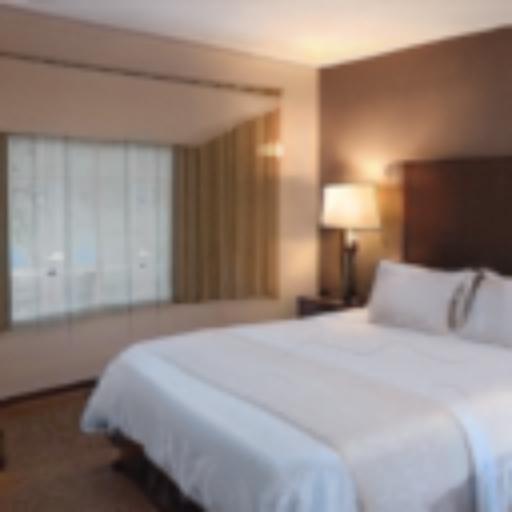}
        \caption{Bilinear}
    \end{subfigure}
    \begin{subfigure}[b]{0.22\linewidth}
        \centering
        \includegraphics[width=\textwidth]{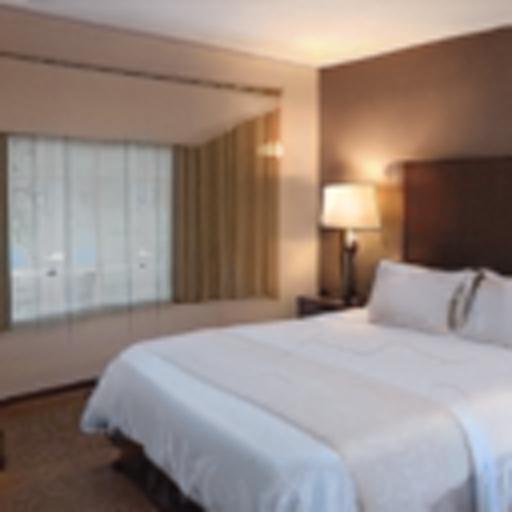}
        \caption{Bicubic}
    \end{subfigure}
    \begin{subfigure}[b]{0.22\linewidth}
        \centering
        \includegraphics[width=\textwidth]{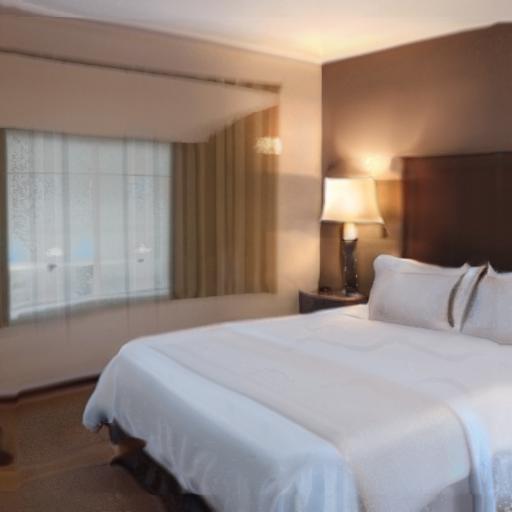}
        \caption{INR-based}
    \end{subfigure}
\end{subfigure}

    \caption{Additional samples from our model to show superresolution properties. We trained the model on LSUN $128\times 128$ and upsampled to $256 \times 256$.}
    \label{fig:superresolution:extra}
\end{figure*}

\begin{figure*}
    \centering
    \includegraphics[width=0.9\linewidth]{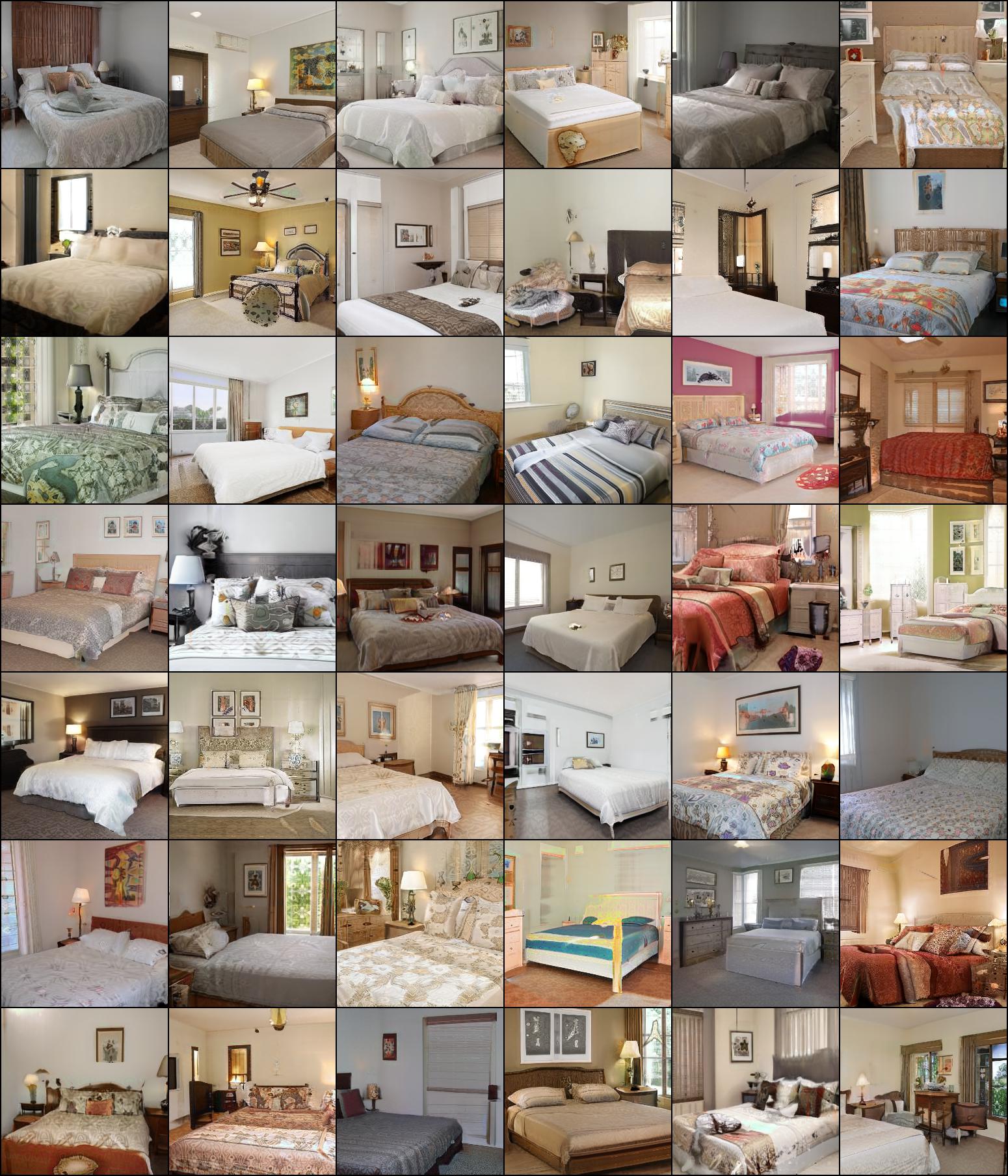}
    \caption{Random samples of our model on LSUN bedroom $256^2$ dataset. FID: 6.27.}
    \label{fig:lsun256}
\end{figure*}

On \figref{fig:ffhq1024-samples}, we present additional samples with the truncation factor of 0.9 from our INR-based model trained on FFHQ1024.
We perform the truncation in similar nature to StyleGAN2 \cite{StyleGAN2} by linearly interpolating an inner representation inside $\G$ to its averaged value.
On \figref{fig:ffhq1024-artifacts}, we present common artifacts found in the produced images.
On \figref{fig:superresolution:extra}, we present additional superresolution samples from our model trained on LSUN $128^2$.
On \figref{fig:lsun256}, we present additional uncurated samples of our model trained on LSUN bedroom $256^2$.

\section{Positional encoding of coordinates}

Recent works \cite{SIREN, FourierINR} demonstrate that using positional embeddings \cite{Transformer} like Fourier features greatly increases the expressivity of a model, allowing to fit more complex data.
Our positional encoding of coordinates follows \cite{FourierINR} design and consists on a linear matrix $\bm W \in \R^{n \times 2}$ applied to raw coordinates vector $\bm p = (x, y)$ and followed by sine/cosine non-linearities and concatenated:

\begin{equation}
\begin{split}
\bm e(\bm p) = \begin{bmatrix}
\sin(\bm W \bm p) \\
\cos(\bm W \bm p)^\top
\end{bmatrix}
\end{split}
\end{equation}

\begin{figure*}
    \centering
    \includegraphics[width=0.9\textwidth]{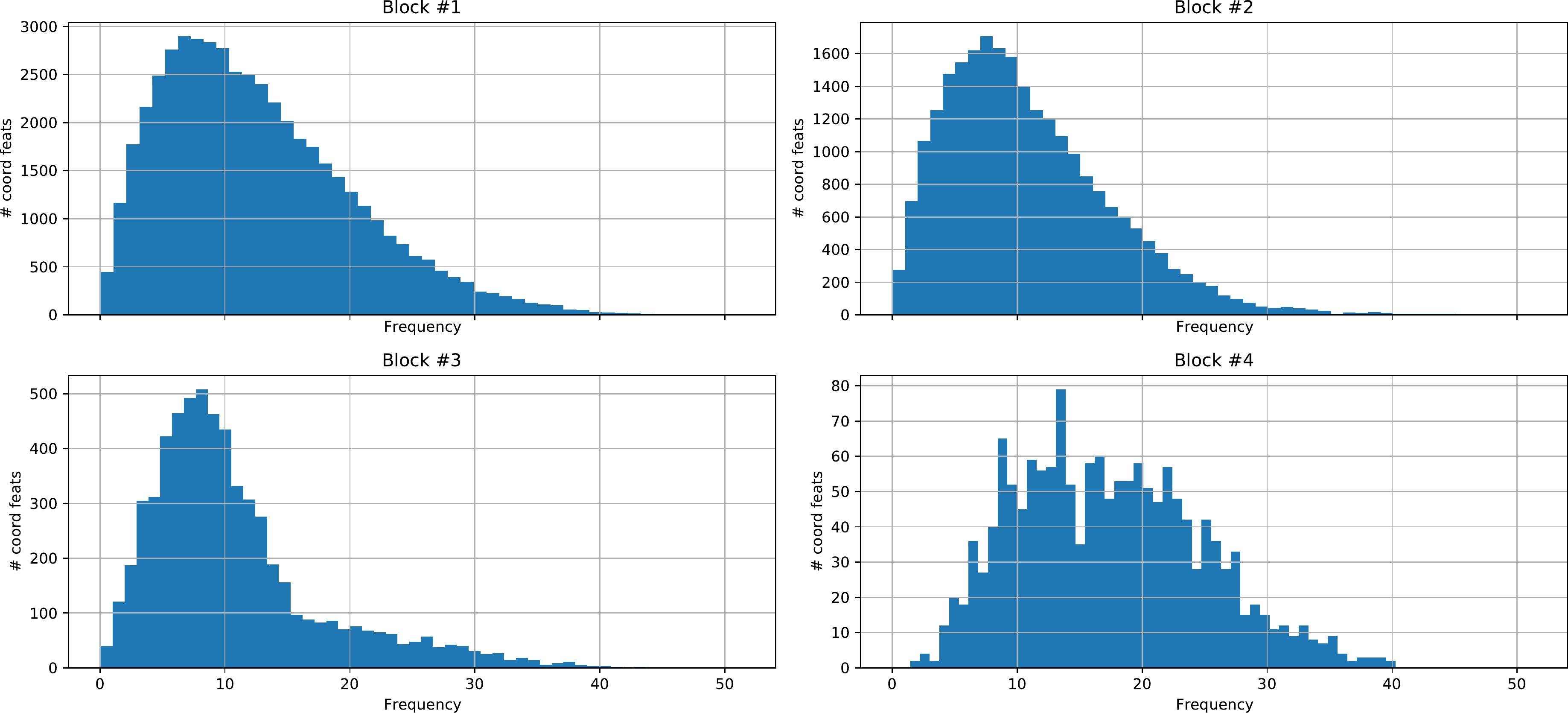}
    \caption{Frequencies distributions for different multi-scale INR blocks of our INR-based GAN trained on FFHQ $1024^2$. To produce the plot, we sample 128 images in an INR-based form and computed the norms of the positional encoding layers, i.e. those layers which take raw coordinates as an input. As one can see, the model tries to use more high-frequent positional embeddings for the last layer since they are more important for drawing fine-grained details. Early layers determine the structure of an image and hence use smaller frequencies to operate on a larger scale.}
    \label{fig:freqs-dists}
\end{figure*}

Matrix $\bm W$ is produced by our generator $\G$ without any factorization since it has only 2 columns.
Each row of this matrix corresponds to the parameters of the Fourier transform.
The norm of a row corresponds to the frequency of the corresponding wave.
We depict frequencies distribution learned by our generator on Figure~\ref{fig:freqs-dists}.

\section{Geometric prior}
Adding coordinates to network input induces powerful prior on the geometric shapes, since now different pixels, otherwise created equal are ordered through the euclidean (or any other) coordinate system. It is a well-known fact that complex geometric shapes could be compactly represented with the use of euclidean coordinates (for example one can write down an ellipse equation as $\frac{(x-x_0)^2}{a^2} + \frac{(y-y_0)^2}{b^2}=1$ ). On the other hand without any form of prior one would hope to fit complex patterns with dedicated filters which would potentially consume much more parameters. 

It is worthy to discuss the synergy between the coordinate representations and hypernetworks. Lets imagine a learnable system consisting of sequentially connected linear layer $W \in \mathbb{R}^{2 \times 2}$, which is modulated by a hypernetwork, and INR, taking Euclidean coordinates as an input $f(X), X \in \mathbb{R}^{2\times1}$ The whole model then could be written as $f(W(z) \times X)$. Now, it could be seen, that introducing the hypernetwork to the pipeline allows to apply linear transformation to the coordinates, rotating and zooming the image encoded by INR $f$. Thus, hypernetworks allow to easily perform transformations non-trivial for traditional deep learning systems.

\noindent \textbf{Finer control over form and placement}
Recent studies show that convolutional NN struggle with such simple and crucial tasks as accurately predicting the coordinates of a drawn point \cite{CoordConv} (and vice versa, drawing a point given the coordinates). One should expect, that such an important skill is necessary for the generative model for accurate placement of the different object parts and for precise representation of the object proportions.
\section{FMM as a generalizaton of the common weight modulation schemes}
In this section we show that Factorized Matrix Multiplication could be seen as a general framework for weight modulation, with Squeeze-and-Excitation, AdaIN and "vanilla" hypernetworks as its particular cases. 

\noindent\textbf{Squeeze-and-Excitation} Let us look at the $l$-th FMM layer of our network with the effective rank of 1. In this case $\bm A^l$ and $\bm B^l$ are matrices (actually vectors) of the sizes $n_{in} \times 1$ and $1 \times n_{out}$ respectively. Thus, following the rules of matrix multiplication, we get that $\bm W_{h \, i,j}^l = \bm A^l_i \cdot \bm B^l_j$. On the other hand, lets look at the Squeeze-and-Excitation mechanism. Here we are modulating the output of the each neuron by multiplicating it by the predicted coefficient, which is equivalent to the multiplication of the corresponding weight matrix column by this coefficient. Using our notions and denoting the pre-activation (before non-linearity) vector of modulation coefficients as $\bm A$ we get that in this case $\bm W_{h \, i,j}^l = \bm A^l_i$. While at the first glance it looks like our model is more expressive lets not forget about the fact that the next layer is by itself modulated with its own Squeeze-and-Excitation, from which (omitting relu non-linearity and the fact that it has its own sigmoid) we can get the $\bm B^l_j$ multiplier. Thus it could be seen that squeeze-and-excitation modulation is roughly equivalent to the FMM layer with the rank of 1. The same reasoning is applicable to the AdaIN case, though, with AdaIN obviously we have the additional normalization layer and do not have sigmoid non-linearity for the style vector which influences the learning dynamics in its own way. We can say that squeeze-and-excitation is the least powerful weight modulation scheme, which uses the matrix of the rank 1 to modulate the main shared weights, on the other hand it is cheap and simple. 

\noindent\textbf{Hypernetworks} Vanilla hypernetwork is perhaps the most straightforward (and the most expensive but flexible) approach to the weight modulation. It is as simple as predicting each weight as an output of MLP. So let's demonstrate that any weight dynamic that can be modeled by hypernetwork could be fitted with the FMM of high enough rank. Let's assume that $n_{in}$ is larger than $n_{out}$ (which is our case, but not essential for the generality of the proof) and choose the FMM rank of $n_{in}$. In this case $\bm A^l$ and $\bm B^l$ are matrices of the sizes $n_{in} \times n_{in}$ and $n_{in} \times n_{out}$ respectively. Since $A$ is a square matrix we can set it to identity constant (which is a solution easily learnt by a NN just by setting bias) and get $\bm W_{h \, i,j}^l = \bm B^l_{i,j}$. In this case any hypernetwork could be "simulated" with FMM by fitting $\bm B^l_{i,j} = \sigma^{-1} (\frac{\bm{\hat{W}}^l_{i,j}}{\bm{W}^l_{s \, i,j}})$, where $\bm{\hat{W}}$ denotes the weight matrix predicted by the hypernetwork. While $\sigma^{-1}$ definitely imposes some restrictions, caused by the positiveness and the range, in our experiments adding sigmoid has not resulted in any harm, perhaps because of the flexible calibration of the $\bm{W}^l_{s}$.

We have shown that main weight modulation schemes could be seen as the boundary particular cases of our approach. Our approach to the weight modulation is somewhere in between of these two extremes, while reaping the benefits of the both of them. Studying the behaviour of this transmission is specially important for the shading light on the weight modulation at whole and going beyond straightforward approaches.
\section{Performance on multi-class datasets}\label{ap:diverse-datasets}

In this section, we conduct experiments on two diverse datasets to demonstrate that our proposed architectural design improves performance in this scenario is well.
For this, we employ two datasets: LSUN-10 $256^2$ and MiniImageNet-100 $128^2$.
LSUN-10 consists on 1M images of 10 LSUN scenes, where we take 100k images of each scene.
MiniImageNet-100 consists on 100k images of 100 ImageNet classes\footnote{\href{https://github.com/yaoyao-liu/mini-imagenet-tools}{https://github.com/yaoyao-liu/mini-imagenet-tools}}, where each class provides 1k images.
We report the results for different models in Table~\ref{table:diverse-datasets}.
They demonstrate that our proposed architectural design improves the performance for this setup as well.

\begin{table}
\caption{FID \& IS at 300k iterations on multi-class datasets.}
\label{table:diverse-datasets}
\centering
\begin{tabular}{|l|cc|cc|}
\hline
\multirow{2}{*}{Decoder type} & \multicolumn{2}{c|}{LSUN-10} & \multicolumn{2}{c|}{MiniImageNet} \\
& FID $\downarrow$ & IS $\uparrow$ & FID $\downarrow$ & IS $\uparrow$ \\
\hline
Basic INR decoder & 216.8 & 1.0 & 271.5 & 1.03 \\
+~Hypernetwork-based decoder & OOM & OOM & 112.9 & 8.76 \\
+~Fourier embeddings & OOM & OOM & 102.8 & 9.85 \\
+~FMM & 23.78 & 2.48 & 84.66 & 9.32 \\
+~Multi-scale INR & 12.47 & 3.02 & 59.63 & 11.29 \\
\hline
StyleGAN2 & 8.99 & 3.18 & 52.94 & 12.32 \\
\hline
Validation set & 0.42 & 9.93 & 0.39 & 61.79 \\
\hline
\end{tabular}
\end{table}

\end{document}